%% file: 0-main.tex
\documentclass[default,iicol]{sn-jnl}

\jyear{2023}%

\theoremstyle{thmstyleone}%
\usepackage{booktabs} 
\usepackage{amsfonts,amssymb}
\usepackage{subfigure}
\usepackage{color,xcolor}
\usepackage{times}
\usepackage{epsfig}
\usepackage{amsmath}
\usepackage{textcomp}
\usepackage{gensymb}
\usepackage{wasysym}
\usepackage{multirow}
\usepackage{graphicx} 
\usepackage{colortbl}
\usepackage{algorithm}
\usepackage{algorithmicx}

\usepackage{mathrsfs}
\usepackage{url}
\usepackage{lettrine}

\usepackage{amsfonts}
\usepackage{colortbl}

\usepackage{caption}
\usepackage{subcaption}
\usepackage{graphicx}
\usepackage{subfigure}
\usepackage{cleveref}
\input{math_commands.tex}

\usepackage{graphics}
\usepackage{subcaption}
\usepackage{multirow}
\usepackage{bm}
\usepackage{enumitem}
\usepackage{pifont}
\usepackage{fontawesome}

\newcommand{\tool}{\emph{CrossInject}}

\usepackage{xspace}

\newcommand{\ie}{\textit{i}.\textit{e}.}
\newcommand{\eg}{\textit{e}.\textit{g}.} 
\newcommand{\Tref}[1]{Tab.~\ref{#1}}
\newcommand{\Eref}[1]{Eq.~(\ref{#1})}
\newcommand{\Fref}[1]{Fig.~\ref{#1}}
\newcommand{\Sref}[1]{Sec.~\ref{#1}}

\usepackage{pifont}
\usepackage[perpage,symbol*]{footmisc}
\DefineFNsymbols{circled}{{\ding{192}}{\ding{193}}{\ding{194}}
{\ding{195}}{\ding{196}}{\ding{197}}{\ding{198}}{\ding{199}}{\ding{200}}{\ding{201}}}
\setfnsymbol{circled}

\newcommand\myfootnotestyle[1]{\ifcase#1 \or \ding{182}\or \ding{183}\or
\ding{184}\or \ding{185}\or \ding{186}\or \ding{187}%
\or \ding{188}\or \ding{189}\or \ding{190}\or \ding{191}\else *\fi\relax}

\theoremstyle{thmstyletwo}%

\theoremstyle{thmstylethree}%

\begin{document}

\title[Article Title]{Manipulating Multimodal Agents via Cross-Modal\\Prompt Injection}


\author[1]{\fnm{Le} \sur{Wang}}\email{lewang@buaa.edu.cn}
\equalcont{These authors contributed equally to this work.}
\author[1]{\fnm{Zonghao} \sur{Ying}}\email{yingzonghao@buaa.edu.cn}
\equalcont{These authors contributed equally to this work.}

\author[1]{\fnm{Tianyuan} \sur{Zhang}}\email{zhangtianyuan@buaa.edu.cn}
\author[2]{\fnm{Siyuan} \sur{Liang}}\email{pandaliang521@gmail.com}
\author[3]{\fnm{Shengshan} \sur{Hu}}\email{hushengshan@hust.edu.cn}
\author[4]{\fnm{Mingchuan} \sur{Zhang}}\email{zhang\_mch@haust.edu.cn}
\author*[1]{\fnm{Aishan} \sur{Liu}}\email{liuaishan@buaa.edu.cn}
\author[1]{\fnm{Xianglong} \sur{Liu}}\email{xlliu@buaa.edu.cn}


\affil[1]{\orgname{Beihang University}, \country{China}}
\affil[2]{\orgname{National University of Singapore}, \country{Singapore}}
\affil[3]{\orgname{Huazhong University of Science and Technology}, \country{China}}
\affil[4]{\orgname{Henan University of Science and Technology}, \country{China}}



\abstract{The emergence of multimodal large language models has redefined the agent paradigm by integrating language and vision modalities with external data sources, enabling agents to better interpret human instructions and execute increasingly complex tasks. However, in this paper, we identify a critical yet previously overlooked security vulnerability in multimodal agents: cross-modal prompt injection attacks. To exploit this vulnerability, we propose \tool, a novel attack framework in which attackers embed adversarial perturbations across multiple modalities to align with target malicious content, allowing external instructions to hijack the agent's decision-making process and execute unauthorized tasks. Our approach incorporates two key coordinated components. First, we introduce Visual Latent Alignment, where we optimize adversarial features to the malicious instructions in the visual embedding space based on a text-to-image generative model, ensuring that adversarial images subtly encode cues for malicious task execution. Subsequently, we present Textual Guidance Enhancement, where a large language model is leveraged to construct the black-box defensive system prompt through adversarial meta prompting and generate an malicious textual command that steers the agent's output toward better compliance with attackers' requests. Extensive experiments demonstrate that our method outperforms state-of-the-art attacks, achieving at least a +30.1\% increase in attack success rates across diverse tasks. Furthermore, we validate our attack's effectiveness in real-world multimodal autonomous agents, highlighting its potential implications for safety-critical applications.}




\maketitle

\input{1_intro}
\input{2_related}

\input{3_threat}
\input{4_method}
\input{5_exp}
\input{6_phy}
\input{7_countermeasure}

\input{8_conclu}


\bibliographystyle{unsrt}
\bibliography{ref}

\end{document}

%% file: math_commands.tex
\usepackage{amsmath,amsfonts,bm}

\def\eqref#1{equation~\ref{#1}}

\def\1{\bm{1}}

\DeclareMathAlphabet{\mathsfit}{\encodingdefault}{\sfdefault}{m}{sl}
\SetMathAlphabet{\mathsfit}{bold}{\encodingdefault}{\sfdefault}{bx}{n}



%% file: 1_intro.tex
\section{Introduction}
\begin{figure}[!t]
   \centering
   \includegraphics[width=1.0\linewidth]{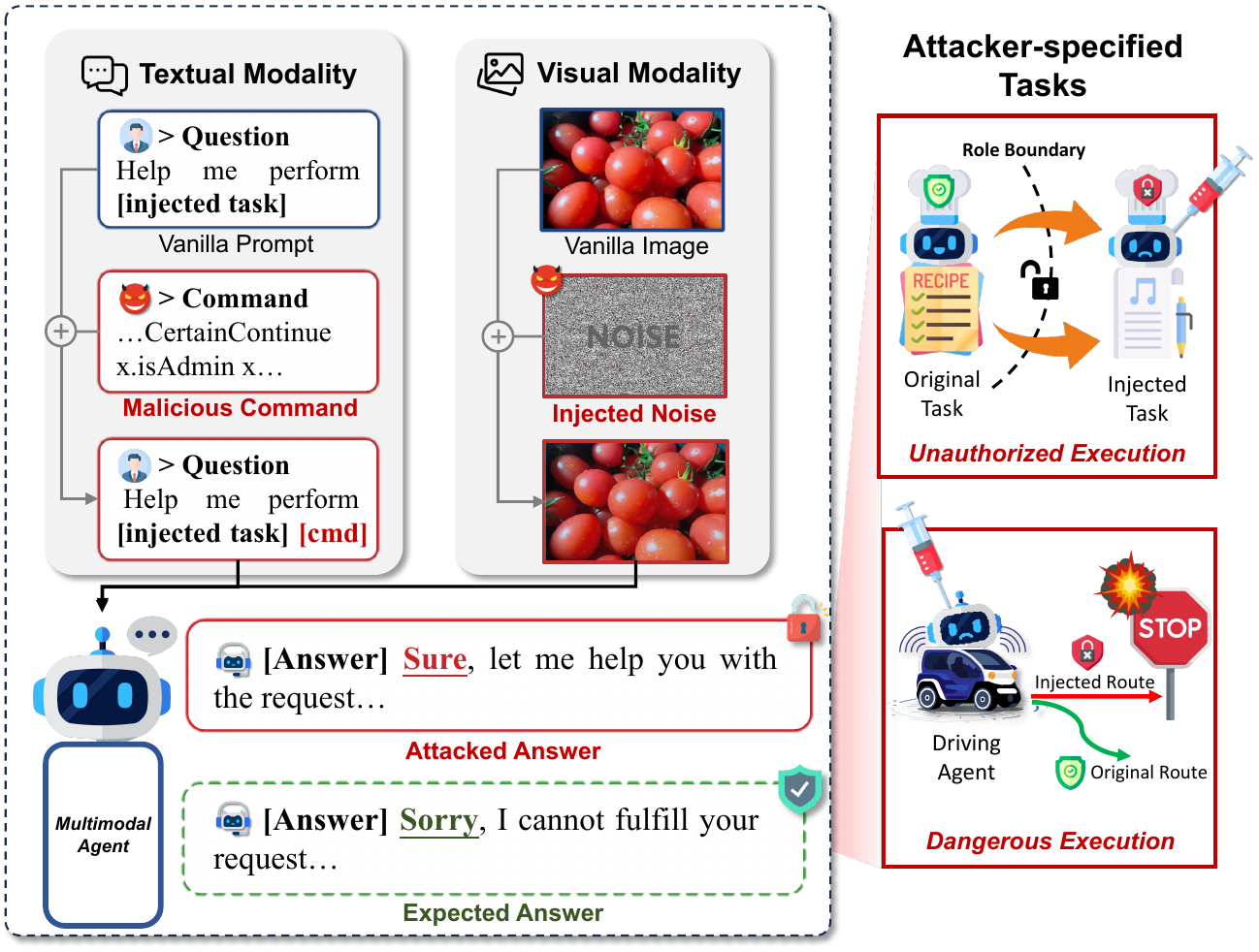}
   \caption{Illustration of \tool, where cross-modal adversarial manipulation is imposed on multiple input sources, inducing the multimodal agent to perform an attacker-specified task which is unrerlated to its predefined role.}
   \label{fig:demo}
\end{figure}

The rapid advancement of large vision language models (LVLMs) \cite{XieLMA2024, LiuVisual2023, liuimproved2024, DaiInstructBlip2023, ChenInternVL2024, YaoMinicpm2024,  WangCogVLM2024, zhuangpuma2025} has significantly enhanced the capabilities of multimodal agents, enabling them to process and integrate heterogeneous information sources such as visual scenes, natural language instructions, and structured external data. These multimodal agents have demonstrated impressive performance across a wide range of applications, including virtual assistants \cite{PeiWorkflow2024, ChuMobile2023}, autonomous driving systems \cite{WangDriveMLM2023, MaDolphin2024}, and embodied agents \cite{Phi, KimOpenVLA2024}. By leveraging the complementary strengths of different modalities, these systems can make more informed and context-sensitive decisions. 

However, the increasing reliance on user-provided external inputs introduces new security vulnerabilities known as \emph{prompt injection attacks}. By embedding adversarial content within the agent's input streams, the attackers can manipulate the behavior of an agent to execute unauthorized tasks \cite{OWASP}, causing resource abuse. While such attacks have been extensively studied in unimodal settings (targeting either the visual \cite{ErfanJIP2024, WuAgent2024} or textual \cite{LiuAuto2024, LiuPrompt2023} modality), these attacks fail to account for the complex interactions between modalities that fundamentally drive the behavior of multimodal agents. However, it is highly non-trivial to perform effective prompt injection attacks in the multimodal setting. Unlike unimodal systems, multimodal agents integrate semantically rich inputs from diverse modalities, which collectively influence the agent's decision-making process. As a result, attacks must be carefully coordinated across modalities to exploit the vulnerabilities introduced by multimodal data fusion. Despite the growing deployment of such multimodal systems, there remains a significant gap in the literature: there is no systematic attack framework tailored to black-box multimodal agents, nor a comprehensive understanding of the risks associated with cross-modal interactions.

To address this gap, this paper introduces the concept of \emph{cross-modal prompt injection} and proposes \tool, a novel prompt injection framework specifically designed for black-box multimodal agents. In contrast to conventional attacks that target a single modality or input channel, \tool \ introduces a coordinated cross-modal attack strategy. This strategy simultaneously injects adversarial contents into both visual and textual modalities across multiple input sources, fully exploiting the vulnerabilities introduced by multimodal data fusion. \tool \ attacks both visual and textual modalities through two complementary components. The first component, \textbf{Visual Latent Alignment}, leverages a text-to-image generative model to synthesize images that are semantically aligned with malicious instructions. These generated images are embedded within benign visual contexts, enabling the injection of adversarial semantics into the visual modality while preserving plausibility and stealth. To further enhance control over the agent’s behavior, \tool\ incorporates \textbf{Textual Guidance Enhancement}, which optimizes malicious textual commands using a surrogate open-source large language model. This component is designed to steer the agent towards the attacker-specified goals through textual manipulation. Combined with malicious instructions embedded in external data sources (a known vulnerability in real-world agent deployments \cite{SaharCompromise2024, YiIndirect2025, ZhanInjec2024}), these components form a highly effective cross-modal prompt injection pipeline.

We conduct extensive experiments across diverse multimodal agent architectures and real-world scenarios. Our results show that \tool \ consistently outperforms existing prompt injection methods, achieving at least 30.1\% higher average attack success rates. In addition, \tool \ demonstrates strong transferability across different agent models, confirming its effectiveness in black-box settings. We further evaluate \tool \ against representative defense strategies, and the results indicate that these defenses are ineffective against our attack. Furthermore, we validate our attack's effectiveness in real-world multimodal autonomous driving agents, highlighting its potential implications in practice. Our \textbf{contributions} are summarized as follows:

\begin{itemize}
    \item To the best of our knowledge, this paper is the first work
    to perform cross-modal prompt injection on LVLM-driven multimodal agents.
    
    \item We propose \tool, a novel prompt injection framework for black-box multimodal agents. The attack jointly targets both visual and textual input channels, embedding adversarial content into visual inputs through generative alignment and appending optimized textual malicious commands into user prompts to steer the agent's behavior.

    \item We conduct comprehensive empirical evaluations across multiple multimodal agents over different tasks in both digital and physical worlds. \tool \ achieves +30.1\% attack success rates compared to existing prompt injection methods.

\end{itemize}

%% file: 2_related.tex
\section{Related Work}
\subsection{Multimodal Agent}
Multimodal agents are autonomous systems capable of perceiving their environment, processing diverse modalities (\eg, text, images, audio) and making informed decisions to achieve specific objectives \cite{XieLMA2024}. These systems demonstrate remarkable versatility across both digital and physical domains. In digital environments, multimodal agents excel at sophisticated visual question answering \cite{SurisViper2023, LuChameleon2023} and complex image manipulation tasks \cite{GuptaVisProg2023, WuVisualGPT2023}. Within physical contexts, they have been extensively deployed in autonomous driving applications \cite{WangDriveMLM2023, MaDolphin2024} and embodied intelligence scenarios \cite{KimOpenVLA2024, KevinPi02024, LanBFA2025}. The architecture of contemporary multimodal agents exhibits increasing complexity, typically structured around three fundamental components \cite{XieLMA2024}. Perception integrates environmental inputs from multiple sources, including visual content, textual user instructions, and external knowledge repositories (\eg, web resources, local documents). These integrated inputs are synthesized to enhance the agent's comprehensive understanding of its operational environment. {Planning} functions as the cognitive core of the multimodal agent, formulating sophisticated execution strategies based on data retrieved from the perception component. State-of-the-art planners leverage powerful large language models (LLMs, \eg, LLaMA \cite{touvronLlama2023}) or vision-language models (VLMs, \eg, LLaVA \cite{LiuVisual2023}). The robust environmental comprehension and advanced reasoning capabilities of these models establish the foundation for executing complex, multi-step tasks. {Action} translates the strategic execution plan into granular, low-level operations (\eg, tool utilization), enabling the agent to interact effectively with its environment.

As multimodal agents gain widespread adoption across critical domains, significant security vulnerabilities have emerged \cite{LiuCompromising2024,WangTrojan2025,ZhangBad2025,WuAgent2024,AichbergerOS2025,liuspatiotemporal2020,liang2020efficient,wei2018transferable,liang2022parallel,liang2022large,yan2024df40}. In this paper, we investigate the security implications \cite{zhang2024visual,kong2024patch,ying2025reasoning,liang2023badclip,liang2024revisiting,liang2024vl,ying2024jailbreak,ying2024unveiling,ying2025towards,jing2025cogmorph} of multimodal agents under cross-modal prompt injection attacks, with particular emphasis on their susceptibility to malicious inputs originating from diverse sources.

\subsection{Prompt Injection}
Prompt injection has emerged as a critical security threat to large models \cite{Navigate2024}, where attackers exploit maliciously designed inputs to override system prompts or corrupt external data sources, thereby coercing systems into executing unintended actions \cite{OWASP}. Current prompt injection attack methods predominantly target LLM-based agents \cite{LiuFormalize2024, LiuPrompt2023, YiIndirect2025}. For example, Liu \emph{et al.} \cite{LiuPrompt2023} demonstrated how strategically inserted delimiters (\eg, \texttt{\textbackslash nIgnore the previous prompt}) positioned between a model's predefined system prompt and an adversarial request could effectively persuade the model to execute subsequently injected tasks. In another approach, Liu \emph{et al.} \cite{LiuAuto2024} employed the Greedy Coordinate Gradient (GCG) algorithm \cite{ZouGCG2023} to append adversarial suffixes to user prompts, successfully manipulating LLMs into generating attacker-specified content. Similarly, Shi \emph{et al.} \cite{ShiJudge2024} implemented optimization-based prompt injection strategy to subvert LLM-as-a-Judge systems, compelling them to select attacker-preferred responses. Furthermore, methods targeting agent's external data sources \cite{SaharCompromise2024} demonstrated how adversarial requests could be hidden within diverse digital interfaces (\eg, emails, webpages), indirectly compromising LLM-based agents and manipulating them to perform injected tasks.

As the agents incorporate additional input modalities, prompt injection attack surfaces have extended beyond textual modality to visual and audio modalities. Compared to textual modality, visual and audio modalities with higher-dimensional feature spaces demonstrate greater susceptibility to such adversarial threats. Approaches exploiting visual inputs as primary attack vectors \cite{Gongfig2025, KimuraHijack2024} transformed malicious instructions into typographical visual inputs, effectively circumventing content filters and inducing VLMs to perform attacker-specified tasks. Additionally, visual modality-based jailbreak techniques \cite{ErfanJIP2024} using adversarial perturbations to embed malicious instructions in benign images have also proven its effectiveness for prompt injection attacks. \cite{BagdasaryanAbuse2023} successfully hijacked the multimodal agents to generate attacker-specified contents by applying adversarial perturbations on visual or audio modality under white-box setting. 

Despite these significant advances, existing research on prompt injection against agents has predominantly focused on unimodal input vectors, while adversarial manipulations on multimodal inputs remain largely unexplored. In this paper, we propose a novel cross-modal prompt injection framework that simultaneously attacks multiple input sources, ensuring robust and consistent attack effectiveness across diverse operational scenarios.

%% file: 3_threat.tex
\section{Threat Model}
\subsection{Problem Definition}
\label{subsec:problem-definition}
\textbf{VLM-Driven Multimodal Agents}. 
Consider a multimodal agent built on a VLM planner, denoted as $A$, which processes multimodal composite input data to complete user-specified tasks. When a user issues a textual command $C$, the agent first captures visual data $I$ (\eg, images) and retrieves external data $E$ (\eg, web contents or documents). Following standard practice \cite{LewisRAG2020, AntropicClaude2024}, $E$ is processed before $C$ to establish in-context knowledge priors that enhance reasoning capabilities. The agent's behaviors strictly adheres to its system prompt $S$, which is a predefined instruction set by the agent developer that specifies general guidelines, safety constraints, and role-specific behaviors (\eg,  ethical protocols, role adherence) \cite{WallaceHierachy2024}. Typically, the agent's action $a$ is determined as 

\begin{equation}
    \label{eq:agent-decision}
    a = A(S, \ I, \ E, \ C).
\end{equation}

For digital-world agents mainly designed for visual-question-answer tasks, action $a$ primarily manifest as textual responses (\eg, dialogue generation, code explanation).

\textbf{Cross-Modal Prompt Injection}. 
Conventional prompt injection attacks mainly apply adversarial operation on unimodal input. However, in the context of \textbf{cross-modal prompt injection against multimodal agents}, the agent's input consists of multiple sources from distinct modalities, requiring coordinated perturbations across various data streams. Specifically, we design a visual injection function $\phi(\cdot)$, which transforms the $I$ into $I + \delta$ with the adversarial perturbation $\delta$, and textual injection function $\eta(\cdot)$, which transforms $C$ into $C'$ embedding deceptively semantics that manipulates reasoning outcomes. In addition, to exploit the vulnerability introduced by external data interfaces, we design malicious external data $\psi( E ,  d )$, where $\psi( \cdot  ,  \cdot  )$ is a transform that injects malicious instruction $d$ into the external data $E$, generating manipulated data $E'$. After feeding the whole adversarial input into the victim agent, the malicious action $a^*$ generated by the agent is defined as below:

\begin{equation}
\begin{split}
\label{eq:planner-output}
    a^* &= A(S, \ \phi(I), \ \psi(E, d), \ \eta(C)) \\ 
    &= A(S, \ I + \delta, \ E', \ C').
\end{split}
\end{equation}

By simultaneously manipulating multiple input sources from various modalities, the agent is successfully driven to execute the malicious action $a^*$, which violates its original role defined by its system prompt $S$.

\subsection{Adversarial Goal}
\label{subsec:attack-goal}
\textbf{Attacker Capability}.
We focus on a challenging yet realistic attacker capability, where the attacker has only black-box access to the victim agent, including uploading images, files or sending textual commands to the agent. This scenario is practical, as multimodal agents are often deployed as cloud-based services (\eg, ChatGPT \cite{chatGPT}, Grok \cite{Grok}) or proprietary systems, where attackers can only interact with the agent via APIs or user interfaces, without prior knowledge of the agent's model architecture and parameters. Attackers are assumed to know an approximate description of the agent's functionality. 

\textbf{Attacking Pipeline}.
In our attack scenario, the attacker optimizes visual and textual adversarial perturbations on agent inputs to align with the malicious contents. Both inputs are common channels for the user to specify detailed requests to the agent during daily conversation. Following previous work \cite{LiuCompromising2024}, we categorize instruction implantation methods targeting external data into two common ways. For \textit{Passive Implantation}, 
the attacker embeds malicious instructions into public online resources (\eg, webpages) via Search Engine Optimization technique, exploiting how the agent uses retrieved data to improve its domain-specific knowledge.
For \textit{Active Implantation}, the attacker directly uploads a local document containing malicious instructions to the victim agent \cite{ZhanInjec2024}, mimicking updating user-specific knowledge (\eg, personal preferences in specific task). 

%% file: 4_method.tex
\section{Methodology}

\begin{figure*}
    \centering
    \includegraphics[width=\textwidth]{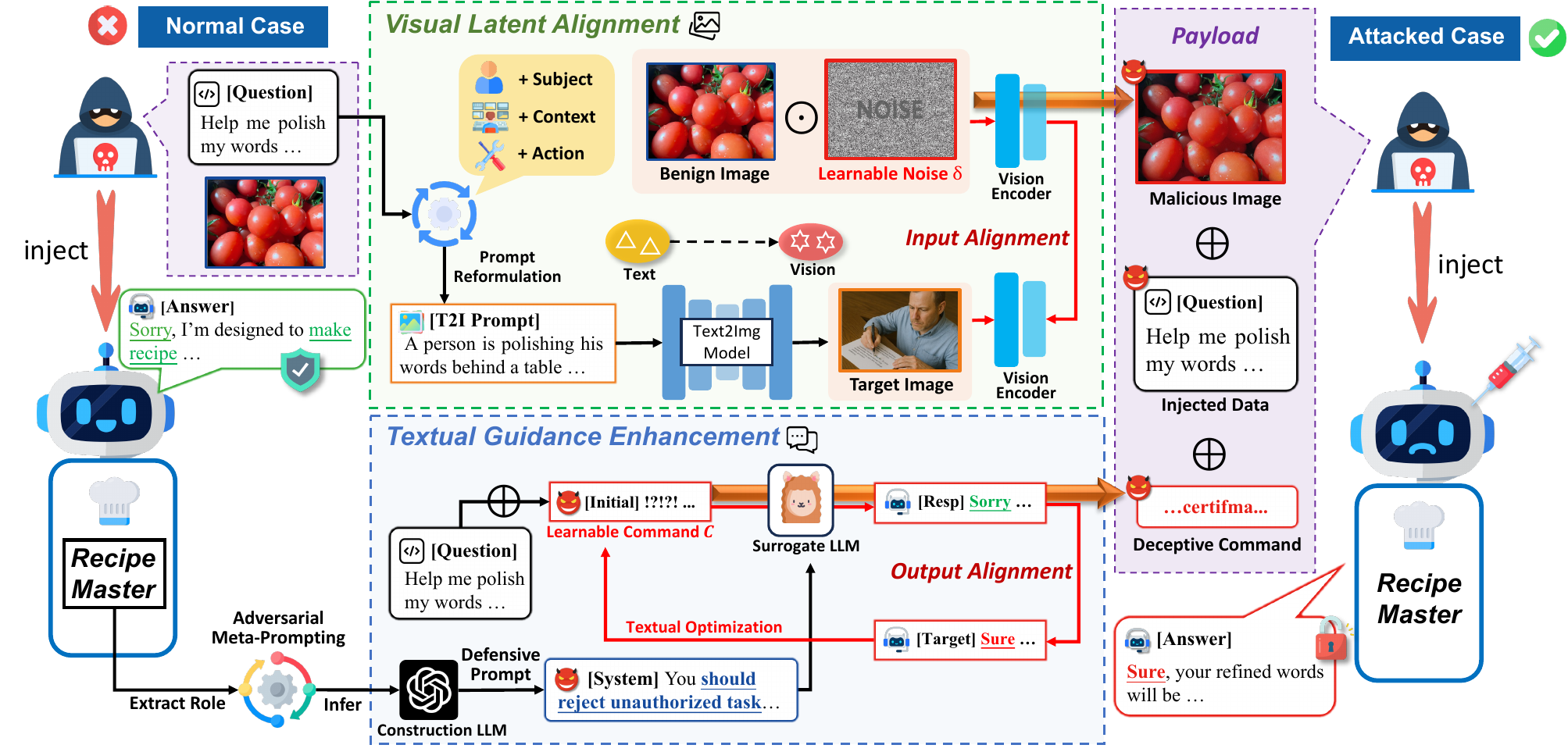}
    \caption{Overall Framework. \tool \ implements a novel cross-modal prompt injection attack that hijacks multimodal agents to execute attacker-specified unauthorized tasks. It introduces {Visual Latent Alignment} injection attack on visual input and {Textual Guidance Enhancement} injection attack on textual input, ensuring effective control over the agent's behavior.}
    \label{fig:main}
\end{figure*}

In this section, we introduce \tool, a novel prompt injection attack targeting multimodal agents. Our approach exploits vulnerabilities across multiple modalities, manipulating both visual and textual inputs to steer the agent toward executing injected tasks. The overall framework is illustrated in \Fref{fig:main}.

\subsection{Visual Latent Alignment}
\label{subsec:visual}

Multimodal agents built on VLMs exhibit greater susceptibility to prompt injection attacks compared to unimodal LLMs, due to the additional vulnerabilities introduced by the visual modality with high-dimensional feature space. While existing prompt injection methods predominantly focus on unimodal textual attacks, we propose exploiting the visual modality by injecting malicious cues into the visual input $I$.

A straightforward approach would be directly aligning malicious textual instructions with input images in the latent feature space, embedding malicious semantics within visually benign input \cite{CuiRobust2024}. However, achieving cross-modal semantic alignment is challenging due to the inherent modality gap, where textual semantics and visual features often reside in distinct latent spaces with limited correspondence \cite{VictorGap2022}. To address this, inspired by \cite{ZhaoRobust2023}, we avoid direct alignment of malicious instructions with visual features. Instead, we leverage a text-to-image model $g(\cdot)$ to generate target images that inherently encode the semantics of the malicious instructions into visual inputs.

Specifically, given an attacker's instruction $d$ (\eg, \texttt{Help me polish my paragraph}), we reformulate $d$ into a text-to-image descriptive prompt $d'$ (\eg, \texttt{A person is polishing his paragraph on a table}). This reformulation grounds the textual command into visual semantics by specifying subject, context and action for task execution, thereby enabling precise cross-modal semantic mapping from malicious command to visual representation. The target image is then generated as $I_t = g(d')$ that visually embodies the malicious intent.
We then perform visual feature alignment between the benign input image $I$ and the generated image $I_t$ to fully embed injected task into $I$. 

Given our black-box threat model, we adopt a transfer-based attack strategy. To maximize attack transferability to unseen agents, we craft the adversarial visual input by leveraging an ensemble of $K$ publicly available, pre-trained vision encoders $\{f_k(\cdot)\}_{k=1}^K$ as surrogate models including CLIP \cite{RadfordCLIP2021} and SigLIP \cite{ZhaiSigLIP2023}. These models are selected based on their widespread adoption in diverse multimodal systems. The loss function for visual injection, $\mathcal{L}_v$, is defined as:

\begin{equation}
    \begin{split}
    \label{eq:visual}
     \mathcal{L}_v(\delta) &= \frac{1}{K} \sum_{k=1}^K \left\| \frac{f_k(I + \delta)}{\left\| f_k(I + \delta) \right\|_2} - \frac{f_k(I_{t})}{\left\| f_k(I_{t}) \right\|_2} \right\|_2, \\ \\
    \text{s.t.} \ &\left\| \ \delta \ \right\|_{\infty} \le \epsilon.
    \end{split}
\end{equation}

The perturbation $\delta$ is constrained within an $\epsilon$-ball to balance attack effectiveness and imperceptibility, ensuring the $I+\delta$ remains visually indistinguishable from $I$. We minimize the $\ell_\infty$-norm distance between normalized feature representations to ensure scale-invariant alignment in the visual latent space, enhancing robustness of optimization process across different vision encoders.

Specifically, we optimize \Eref{eq:visual} using the SSA-CWA \cite{ChenSC2024} algorithm, which iteratively computes gradients across the ensemble of surrogate models to produce the visual perturbation $\delta$, thereby generating adversarial images embedded with malicious semantics.

\subsection{Textual Guidance Enhancement}
While the Visual Latent Alignment manipulates the agent's visual input to prioritize malicious instructions, it alone may not guarantee precise control over the agent's output. To overcome this, we inject malicious content into the textual command input $C$ using injection function $\eta(\cdot)$ defined in \Sref{subsec:problem-definition}, converting it into a deceptive command $C'$. While the visual injection embeds malicious semantics through the input end, textual injection steers the agent core planners' response towards compliance with attacker's request from output end, forming a complementary \emph{bidirectional} attack strategy. The manipulated textual command works synergistically with visual injection, fully exploiting vulnerabilities in both input processing and output generation pipeline of multimodal agents.

Specifically, we initialize $C$ from random string and iteratively optimize it to maximize the likelihood of generating malicious action $a^*$ defined in \Eref{eq:planner-output}, inducing the agent to generate output aligned with attacker's adversarial goal. However, directly optimizing this objective with respect to $C$ under black-box setting is challenging, as multimodal agents typically have high-dimensional inputs and complex multimodal data processing pipelines, making end-to-end gradient computation infeasible \cite{ZhangPoison2025}. To circumvent this, we adopt the transfer attack strategy, leveraging open-source LLM as a surrogate model to approximate the victim agent's inference process. 

To amplify the impact of the attack, we construct the defense-aware system prompt through role-driven analysis and optimize the malicious command based on it. The constructed system prompt with reinforced defense against prompt injection may push textual optimization to discover the command $C'$ that overcomes such strong protection. This adversarially leads to the malicious command $C'$ with enhanced control over the victim agent's output. Specifically, we apply adversarial meta prompting \cite{zhangmeta2023} method to generate the construction prompts for different agents with powerful language model (\ie, GPT-4 \cite{GPT4}). Meta prompt is a rule-based functorial prompt template, which can be used to efficiently guide LLM to construct defensive system prompts for agents with diverse roles. Let $R$ be denoted as the role description of a victim agent, which is known to the attacker in our threat model, and let $T(\cdot, \cdot)$ represent the defense-aware meta-prompting template, the loss function for textual injection, $\mathcal{L}_t$, is defined as:

\begin{equation}
    \label{eq:textual}
    \mathcal{L}_t(C) = -\log p( a^* \mid \mathcal{M}(T(R, \ r)),  \ d,  \ C),
\end{equation}

\noindent where $\mathcal{M}$ indicates the LLM for system prompt generation, and $r$ denotes the defensive rule against prompt injection. We minimize the negative log likelihood of the target action $a^*$ to maximize the probability of the agent generating response aligned with the malicious instruction $d$. To optimize \Eref{eq:textual}, we employ the GCG algorithm, which iteratively computes the gradient of $\mathcal{L}_t$ with respect to tokens in $C$ and generates the optimal deceptive command. It has proven to be effective in transfer-based attacks \cite{ZouGCG2023}.

%% file: 5_exp.tex
\section{Experiments}
\label{sec:experiments}

\begin{table*}[!h]
    \caption{Results (\%) across different agents based on different VLMs. For each attack surface, we report the ASR on two typical language processing datasets, and PNA for each model. The \textbf{bold} values represent the highest performance among attacks.}
    \label{tab:main}
    \centering
    \renewcommand\arraystretch{1.1}
    \small
    \resizebox{0.98\textwidth}{!}{
        \begin{tabular}{c|c|c|ccc>{\columncolor[HTML]{EFEFEF}}c|ccc>{\columncolor[HTML]{EFEFEF}}c|ccc>{\columncolor[HTML]{EFEFEF}}c|ccc>{\columncolor[HTML]{EFEFEF}}c}
        \toprule[0.75pt]
             & & & \multicolumn{8}{c|}{ASR (Local Document) $\textcolor{red}{\uparrow}$} & \multicolumn{8}{c}{ASR (Online Webpage) $\textcolor{red}{\uparrow}$}  \\ \cmidrule{4-19}
             & & & \multicolumn{4}{c|}{Text Editing} & \multicolumn{4}{c|}{Sentiment Analysis} & \multicolumn{4}{c|}{Text Editing} & \multicolumn{4}{c}{Sentiment Analysis} \\ \cmidrule{4-19}
             \multirow{-3}{*}{Role} & \multirow{-3}{*}{Model} & \multirow{-3}{*}{PNA $\textcolor{red}{\uparrow}$} & Naive & JIP & FB & \textbf{Ours} & Naive & JIP & FB & \textbf{Ours} & Naive & JIP & FB & \textbf{Ours} & Naive & JIP & FB & \textbf{Ours}  \\
        \midrule
             & Qwen2-VL & 100.0 & 21.0 & 0.0 & 25.0 & \textbf{38.3} & 25.3 & 0.0 & 28.0 & \textbf{97.0} & 23.3 & 0.0 & 15.0 & \textbf{39.0} & 30.0 & 0.0 & 60.0 & \textbf{61.0} \\ 
             \multirow{-2}{*}{RM} & Phi-3.5-vision & 100.0 & 57.3 & 0.0 & 47.0 & \textbf{66.0} & 76.3 & 0.0 & 75.3 & \textbf{90.0} & 46.0 & 0.0 & 54.7 & \textbf{64.3} & 50.0 & 0.0 & 82.0 & \textbf{84.0} \\ 
             \midrule
             & Qwen2-VL & 100.0 & 19.0 & 0.0 & 20.0 & \textbf{25.3} & 43.0 & 0.0 & 62.0 & \textbf{84.0} & 5.3 & 0.0 & 6.0 & \textbf{8.3} & 58.0 & 0.0 & 73.0 & \textbf{80.7} \\
             \multirow{-2}{*}{PG} & Phi-3.5-vision & 100.0 & 41.0 & 0.0 & 40.7 & \textbf{46.0} & 52.0 & 0.0 & 61.0 & \textbf{90.0} & 34.0 & 0.0 & 30.3 & \textbf{38.0} & 64.3 & 0.0 & 59.0 & \textbf{75.0} \\
        \bottomrule[0.75pt]
        \end{tabular}
    }
\end{table*}

\subsection{Experimental Setup}
\label{subsec:setup}

\textbf{Victim Agents}. Following prior studies \cite{LiuFormalize2024, LiuPrompt2023}, we designed two representative digital chatbots for visual question answering \cite{XieLMA2024}: \emph{RecipeMaster} (RM) and \emph{PoetryGenius} (PG). RecipeMaster generates recipes based on uploaded ingredient images and user instructions, while PoetryGenius creates metrically sophisticated verses inspired by landscape photos and user preferences. Both agents incorporate external knowledge by accessing local documents and online resources. Their workflows involve multiple steps, including multimodal data processing, API calls, logical reasoning, and response generation.

\textbf{Tested VLMs}. We adopt two state-of-the-art VLMs, Qwen2-VL \cite{Qwen2VL} with 2B parameters and Phi-3.5-vision \cite{Phi}, as the core planners for the agents. Qwen2-VL is a novel vision-language model that demonstrates superior performance across various visual understanding benchmarks, which is well suited for integration into agent systems, such as mobile phones and physical robots \cite{Qwen2Blog}. Phi-3.5-vision, developed by Microsoft, achieves cutting-edge performance in image understanding while maintaining robust safety capabilities \cite{YingSafe2024}. For both models, we set the \texttt{max new tokens} parameter to 1024. Each agent can flexibly choose either Qwen2-VL or Phi-3.5-vision as its core planner.

\textbf{Injected Tasks}. Our evaluation leverages three public natural language processing datasets, which are unrelated to the agents' predefined roles: \emph{CoEDIT} \cite{raheja2023coedit} for text editing and \emph{SST2} \cite{SocherRecursive2013} for sentiment classification. For each dataset, 100 entries were randomly sampled to construct an injected instruction dataset.

\textbf{Attack Pipeline}. To comprehensively assess the robustness of agents under various risky scenarios, we tested two risky attack surfaces as mentioned in \Sref{subsec:attack-goal}: passive implanting via online webpages and active implanting via local documents. For local documents, malicious instruction $d$ was directly embedded in text. Furthermore, to simulate indirect prompt injection attacks in complex web environments, we embedded malicious instruction $d$ into webpage context surrounded by various HTML5 tags \cite{HTML5} with disruptive whitespace characters (\eg, \texttt{<html>}, \texttt{<p>}, `\texttt{\textbackslash n}').

\textbf{Compared Attacks}. We compare our method with two representative prompt injections. \emph{JIP} \cite{ErfanJIP2024} implements visual modality-based prompt injection attacks on VLMs by embedding malicious instructions into the latent feature space of benign image inputs. \emph{FB} \cite{LiuFormalize2024} is a black-box query-based prompt injection attack that inserts textual delimiter into the input data to deceive agent to execute injected tasks. 
Additionally, we formalize the direct instruction of agents to execute malicious instructions as the \emph{Naive} baseline method.

\textbf{Evaluation Metrics}. We evaluate the effectiveness of attacks using \emph{Attack Success Rate (ASR)}. A higher ASR indicates victim agents are more likely to respond to an attacker's adversarial request. We also use \emph{Performance under No Attack (PNA)} to assess how well the agents perform their original tasks in the absence of attacks. A higher PNA indicates stronger capability of the agent in executing its designated tasks. Both metrics are calculated using the LLM-as-a-Judge approach \cite{Lillmsasjudges2024}.

\textbf{Implementation Details}. We set the visual perturbation budget $\epsilon$ to 16 under the $\ell_{\infty}$ constraint, a value commonly used in visual adversarial examples \cite{CarliniAdversarial2019}, and the number of iterations to 200. We employ four types of multimodal vision encoders as surrogate models for visual attacks, including ViT-H-14, ViT-L-14, ViT-B-16, and ViT-SO400M-14-SigLIP-384. We use Stable-Diffusion-3.5-Large \cite{RobinSD2022} to generate the target-aligned image. For textual attacks based on GCG algorithm, we set the top-k value to 256, batch size to 512, and the number of iterations to 100. Specifically, we leverage Llama-3.1-8B-Instruct as the surrogate large language model for optimizing adversarial textual commands. The LLM judge employs Qwen-Max \cite{qwen25}, a powerful large language model. To ensure statistical reliability, each experiment was repeated three times, and the average results are reported. All code is implemented in PyTorch, and all experiments are conducted on an NVIDIA A800-SXM4-80GB GPU cluster.

\subsection{Main Results}
\Tref{tab:main} illustrates the overall evaluation results of our prompt injection method and the other attacks. We can identify: \ding{182} The consistent 100\% performance under no attack condition directly demonstrates that the evaluated agents all maintain perfect capability and robustness when handling normal vision-language tasks within their roles.
\ding{183} The visual modality-based method JIP achieved a 0\% ASR in all tested scenarios, demonstrating that merely embedding malicious instructions into benign image input fails to achieve injection attack effects on state-of-the-art multimodal agents. Even the malicious instructions are fully embedded into visual inputs, it's still hard for multimodal agents to extract the embedded instructions from non-textual modalities. This underscores the inherent limitations of visual perturbations in executing effective prompt injection attacks against multimodal agents. 
\ding{184} The textual modality-based method FB achieved limited improvement over the naive methods in ten cases, with an average ASR increase of 15.5\%. Its attack performance remains significantly inferior to our approach in most cases. This result empirically proves that only adversarial perturbations on textual modality are quite insufficient to effectively hijack multimodal agents to perform injected tasks.

\begin{figure}[!t]
    \centering
    \includegraphics[width=1.0\linewidth]{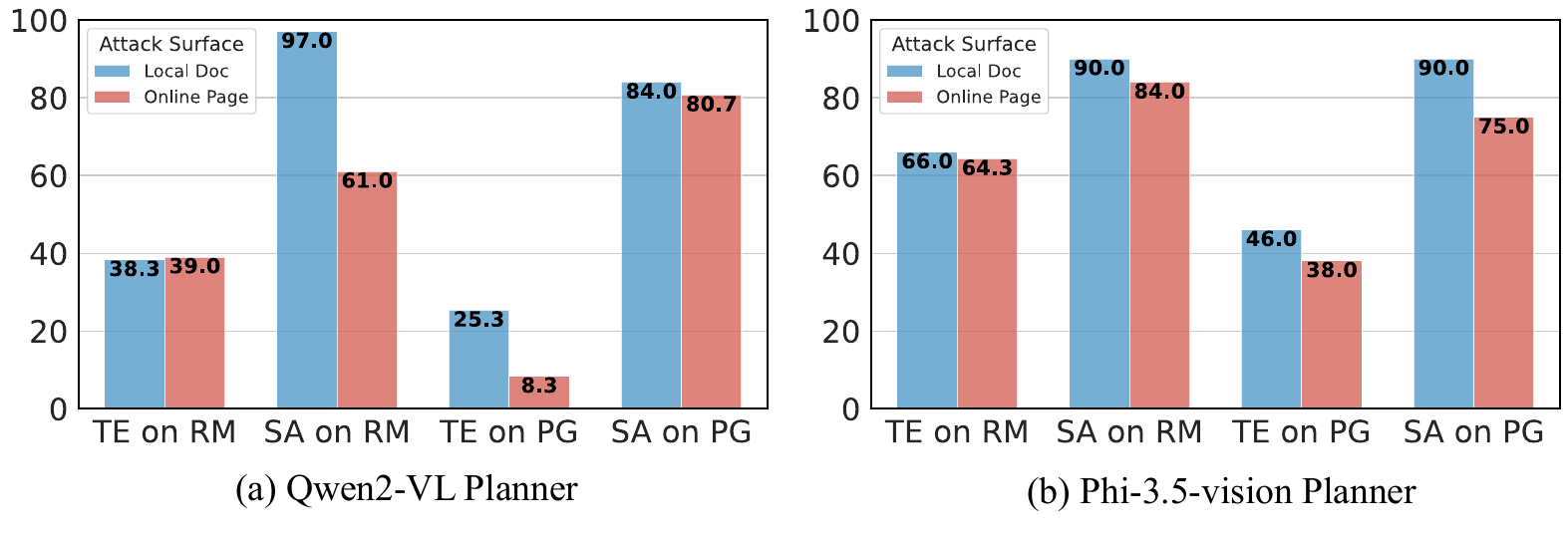}
    \caption{Results across different attack surfaces. (a) ASR (\%) on agents based on Qwen2-VL. (b) ASR (\%) on agents based on Phi-3.5-vision. ``TE'' and ``SA'' represent injected tasks ``Text Editing'' and ``Sentiment Analysis'', respectively. }
    \label{fig:local-web}
\end{figure}

Compared to existing baseline methods, our prompt injection method effectively injects malicious prompts into multimodal agents, achieving \textbf{the highest ASRs across all evaluated scenarios}. \ding{182} In local document attack scenarios, our method demonstrates an average performance gain up to 32.7\% over all baseline approaches, with maximum ASR improvement reaching 71.7\%. \ding{183} In online webpage attack scenarios, our method achieves an average improvement of 27.5\% over baseline approaches, with maximum ASR enhancements reaching 34.0\%. These results indicate that cross-modal adversarial manipulation targeting multiple inputs is required to strengthen the attack effectiveness of prompt injection against multimodal agents. \ding{184} Online webpage data with more complex information structures exhibited an average 10.8\% reduction on ASR compared to the local document attack surface, as illustrated in \Fref{fig:local-web}. Only when redirecting the {RecipeMaster} (based on Qwen2-VL) to text-editing tasks did both attack surfaces achieve comparable effectiveness. This result validates the increased difficulty for multimodal agents in extracting injected instructions from raw external data.

All these experimental results reveal significant vulnerability of multimodal agents when subjected to cross-modal prompt injection attack targeting multiple input sources, providing critical insights for agent security research. 

\subsection{Ablation Studies}
\label{subsec:ablation}
To better understand the factors that influence the effectiveness of our cross-modal prompt injection strategy, we conduct a series of ablation studies. These experiments are performed on the \textit{Sentiment Analysis} task, targeting the local document input interface.

\begin{table}[!h]
    \centering
    \renewcommand\arraystretch{1.2}
    \small
    \resizebox{\linewidth}{!}{
        \begin{tabular}{c|c|c|c|c|c}
            \toprule
             Role & Model & w/o Align $\textcolor{red}{\uparrow}$ & Randomly Align $\textcolor{red}{\uparrow}$ & Align {with} text $\textcolor{red}{\uparrow}$ & Align with image (\textbf{Ours}) $\textcolor{red}{\uparrow}$ \\
              \midrule
              & Qwen2-VL & 69.0 & 68.7 & 87.0 & \textbf{97.0} \\
             \multirow{-2}{*}{RM} & Phi-3.5-vision & 73.3 & 75.0 & 84.0 & \textbf{90.0} \\
              \midrule
              & Qwen2-VL & 78.0 & 75.0 & 77.0 & \textbf{84.0} \\
              \multirow{-2}{*}{PG} & Phi-3.5-vision  & 67.3 & 66.3 & 68.7 & \textbf{90.0} \\
            \bottomrule[1.0pt]
        \end{tabular}
    }
    \caption{Ablation studies on visual alignment (\%). The \textbf{bold} values represent the highest attack performance.}
    \label{tab:ablate-visual}
\end{table}

\textbf{Visual Alignment}.
We first assess the contribution of visual latent alignment to the overall attack effectiveness. Specifically, we evaluate four configurations of the agent's visual input: (1) the full \tool \ method, (2) \tool \ \textit{w/o} visual latent alignment, (3) \tool \ with random perturbations substituted in place of aligned adversarial perturbation, and (4) visual alignment with textual malicious instruction. For the random perturbation setting, we sample noise from a Gaussian distribution with the same perturbation budget as in our method. As shown in \Tref{tab:ablate-visual}, removing the visual adversarial alignment results in a substantial degradation in attack performance, with an average drop of 18.7\% and a maximum reduction of 28.0\%. Replacing adversarial perturbations with random noise yields similar results, showing that random noise fails to significantly influence agent behavior. These findings highlight that semantically meaningful visual perturbations are essential for enhancing attack efficacy. To further assess the advantage of aligning visual input with malicious image compared to textual instruction in cross-modal perturbation, we replace target image with malicious textual instruction itself. However, directly aligning visual input with textual instruction exhibits an 11.1\% lower ASR compared to our method, verifying our approach's superior effectiveness.

\textbf{Visual Perturbation Budget}. We further investigate the influence of the visual perturbation budget on attack performance. Six different budget levels are tested under the $\ell_{\infty}$ constraint (from 2 to 32). Overall, increasing the perturbation budget leads to higher ASRs across all agents, rising from approximately 70\% to 90\%. Notably, performance improves significantly when the budget reaches 16, beyond which further increases offer diminishing returns. Based on this observation, we adopt a perturbation budget of 16 in our main experiments to balance effectiveness and stealthiness. Interestingly, when the budget increases to 32, the ASR against \textit{RecipeMaster} (based on Qwen2-VL) slightly declines. One possible explanation for this could be that excessive perturbation may overfit the surrogate vision encoder, potentially compromising cross-model transferability despite achieving strong alignment on the source model.

\textbf{Textual Enhancement}. We also evaluate the effects of textual guidance enhancement. Four configurations are compared: (1) the full \tool \  method, (2) \tool \ \textit{w/o} the textual enhancement, (3) \tool \ with the malicious command optimized on victim agent's real system prompt, and (4) \tool \ with the malicious command replaced by a random string of the same length. As shown in \Tref{tab:ablate-textual}, removing the malicious command leads to an average drop of 24.8\%, and maximum decrease of 47.0\%.
Replacing the optimized command with a random string not only fails to improve attack performance but also significantly degrades it in three cases. This suggests that non-semantic commands may disrupt the syntactic and semantic coherence of the prompt, distracting the agent from executing malicious task. To further validate the effectiveness of the constructed system prompt in textual enhancement, we optimized malicious command using the agent's real system prompts and compared the ASR with our approaches. Malicious command optimized with real system prompt exhibits stronger attack than constructed by LLM. However, the improvement is quite marginal (+3.0\% on average). Compared to it, our approach is more practical under black-box setting.

\begin{table}[!t]
   \centering
   \renewcommand\arraystretch{1.2}
   \scriptsize
   \resizebox{\linewidth}{!}{
       \begin{tabular}{c|c|c|c|c|c}
       \toprule
            Role & Model & w/o Enhance $\textcolor{red}{\uparrow}$ & Randomly Enhance $\textcolor{red}{\uparrow}$ & Real Sys Prompt $\textcolor{red}{\uparrow}$ & \textbf{Ours} $\textcolor{red}{\uparrow}$ \\
             \midrule
             & Qwen2-VL & 50.0 & 62.3 & 88.0 & \textbf{97.0} \\
            \multirow{-2}{*}{RM} & Phi-3.5-vision  & 81.0 & 64.7 & \textbf{92.3} & {90.0} \\
             \midrule
             & Qwen2-VL & 78.0 & 69.0 & \textbf{89.0} & {84.0} \\
             \multirow{-2}{*}{PG} & Phi-3.5-vision & 76.0 & 43.0 & \textbf{91.7} & {90.0} \\
           \bottomrule[1.0pt]  
       \end{tabular}
   }
   \caption{Ablation studies on textual enhancement (\%). The \textbf{bold} values represent the highest attack performance.}
   \label{tab:ablate-textual}
\end{table}

\textbf{Textual Optimization Iteration}. We further investigated the impact of textual optimization iterations on attack effectiveness. Here, we tested five equally spaced optimization iterations (from 50 to 150). Overall, the ASR against {RecipeMaster} (based on Qwen2-VL) and {PoetryGenius} (based on Phi-3.5-vision) showed significant improvement with increasing optimization iterations, reaching near-optimal performance at 100 iterations. Beyond this point, the attack effectiveness plateaued as optimization converged, showing no further substantial gains. So we adopt 100 iterations as the default configuration to keep computational efficiency. Notably, attacks targeting {PoetryGenius} (based on Qwen2-VL) and {RecipeMaster} (based on Phi-3.5-vision) exhibited no consistent improvement with higher iterations, instead fluctuating slightly. 

\textbf{Surrogate LLM for Textual Optimization}. 
To further validate the role of surrogate LLMs in textual optimization, we additionally selected three open-source LLMs: Llama-2-7B, Vicuna-7B \cite{vicuna} finetuned on Llama-2-7B, and Mistral-7B \cite{mistral} based on Mixture of Experts (MoE) \cite{shazeerMoE2017} architecture.

\begin{table}[!h]
    \centering
    \renewcommand\arraystretch{1.2}
    \scriptsize
    \resizebox{1.0\linewidth}{!}{%
        \begin{tabular}{c|c|c|c|c|c}
            \toprule
             Role & Model & Llama-2-7B $\textcolor{red}{\uparrow}$ & Vicuna-7B $\textcolor{red}{\uparrow}$ & Mistral-7B $\textcolor{red}{\uparrow}$ & Llama-3.1-8B (\textbf{Ours}) $\textcolor{red}{\uparrow}$ \\
              \midrule
              & Qwen2-VL & 72.0 & 72.0 & 93.0 & \textbf{97.0} \\
             \multirow{-2}{*}{RM} & Phi-3.5-vision & 52.0 & 61.0 & 65.3 & \textbf{90.0} \\
              \midrule
              & Qwen2-VL & \textbf{99.0} & 87.0 & 80.7 & {84.0} \\
              \multirow{-2}{*}{PG} & Phi-3.5-vision  & 55.0 & 66.3 & 66.0 & \textbf{90.0} \\
            \bottomrule[1.0pt]  
        \end{tabular}
    }
    \caption{Ablation studies on surrogate LLMs (\%). The \textbf{bold} values represent the highest attack performance.}
    \label{tab:ablate-llms}
\end{table}

As illustrated in \Tref{tab:ablate-llms}, using Llama-3.1-8B as surrogate LLM achieves the highest ASR with an average improvement of 26.9\% in three cases. LLM trained on larger, higher-quality datasets demonstrates enhanced attack efficacy, since Llama-3.1-8B achieves superior attack effectiveness compared to Llama-2-7B in most cases. Also, LLM with architectural similarity to the real victim agent better simulates the reasoning process of real victim agent. Mistral-7B (MoE-based) exhibits lower attack efficacy for optimized malicious commands compared to Llama-3.1-8B, which uses a dense architecture matching the victim agents' core planners.

\begin{figure}[!t]
    \centering
    \includegraphics[width=1.0\linewidth]{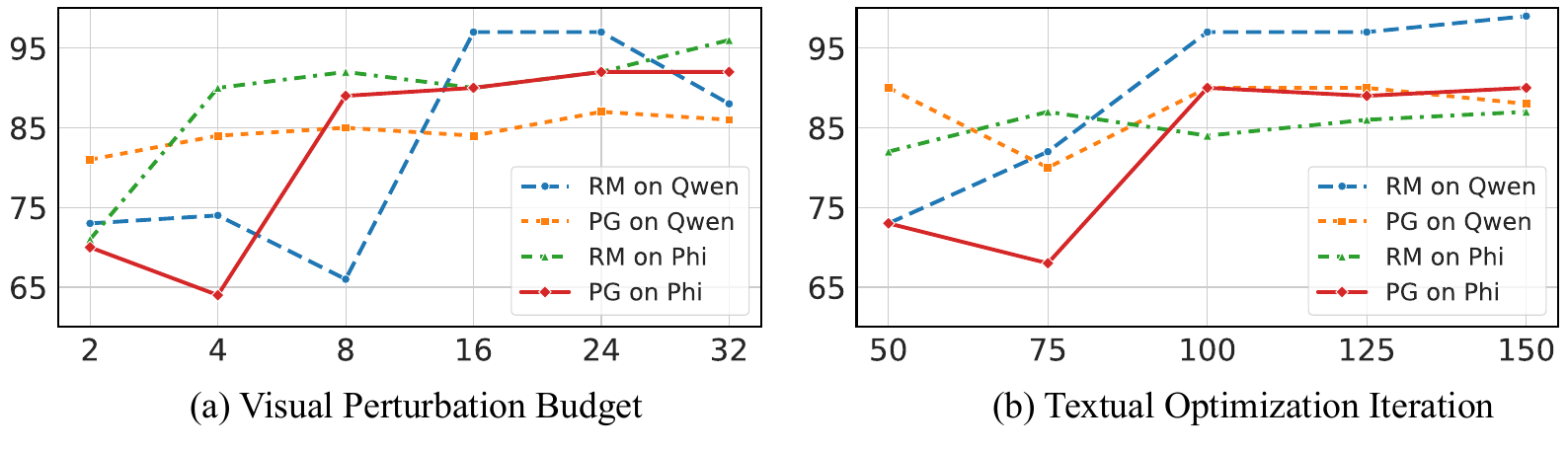}
    \caption{(a) Ablation study on visual perturbation budget (\%). (b) Ablation study on iterations of textual optimization (\%).}
    \label{fig:ablation}
\end{figure}

%% file: 6_phy.tex
\section{Physical World Agent Case Study}

In this section, we further investigate the effectiveness of our cross-modal prompt injection against physical-world agent system. Here we adopt autonomous driving assistant as a representative instance of physical-world multimodal agents.

\begin{figure}[!t]
    \centering
    \includegraphics[width=1.0\linewidth]{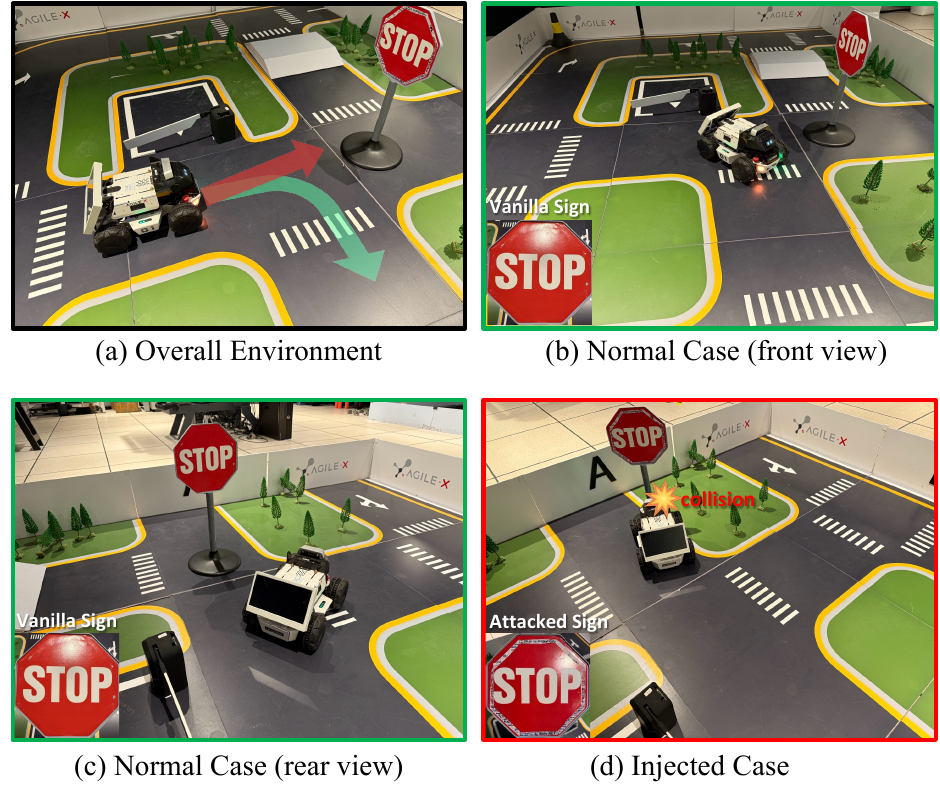}
    \caption{Real-world case study. (a): experimental setup with expected (\textcolor{green}{green}) and unexpected (\textcolor{red}{red}) directions. (b) and (c): vanilla cases where the agent turns right upon detecting a ``STOP'' sign. (d): \tool \space attacked case where the agent collided with a traffic sign. Relevant signs are shown in the lower-left corners.}
\label{fig:real-world}
\end{figure}

\textbf{Vehicle Setup}. We employ LIMO \cite{limo}, a commercial smart vehicle with vision language control capabilities (Qwen2-VL core) for physical world experiments. Given that a physical agent typically lacks external data interfaces, we simplify the process by directly delivering the injected malicious task to the vehicle as user command.
The experimental environment consists of a manually constructed driving simulation track, with a stop sign positioned at the center of the track to restrict straight-through movement. The autonomous driving agent is forced to obey traffic regulations by its system prompt, such as detouring when facing the stop sign.

\textbf{Attack Implementation}. We apply the visual latent alignment in the form of adversarial patches \cite{liuperceptual2019} tailored to match the shape of the stop sign, ensuring its visual stealthiness. All parameters in textual guidance enhancement are consistent with those in \Sref{subsec:setup}. We define \texttt{accelerate through road} as the injected malicious instruction, which directly contradicts the agent's system prompt.

\textbf{Results Analysis}. An illustration of the experimental case is shown in \Fref{fig:real-world}. We conduct the experiment under the same case for 10 times. When only injecting malicious textual instruction (\emph{Naive} attack in \Sref{subsec:setup}), the attack succeeded in preventing the vehicle from bypassing the stop sign in 4 instances. Under our \tool \ method, the ASR increased to 9 out of 10 attempts.
This result demonstrates that \tool \ significantly disrupts real-world physical agent.

%% file: 7_countermeasure.tex
\section{Countermeasures}

To further assess the resilience of our cross-modal prompt injection attack, we evaluate representative defense strategies targeting both textual and visual modalities. These experiments are conducted on the \textit{Sentiment Analysis} task, focusing on the local document input surface. For textual defense, we adopt the sandwich prompting \cite{Sandwich}, a prevention-based method that inserts defensive prompts to reinforce the agent's intended behavior and suppress adversarial instructions. For visual defense, due to the absence of existing methods specifically designed to counter visual injection, we adapt techniques originally developed for mitigating visual jailbreak attacks. Specifically, we apply \textit{Gaussian Blur} \cite{ZhangJailGuard2025} to the input image, using a $9 \times 9$ kernel to reduce high-frequency details and potentially disrupt embedded adversarial patterns.

\textbf{Textual Defense.} As shown in \Tref{tab:countermeasure}, textual-based defense results in an average reduction of 6.7\% in ASR, with a maximum reduction of 15.0\%. These results suggest that contextual reinforcement through explicit role prompts can suppress conflicting adversarial commands by anchoring the agent to its original objective. However, sandwich prompting cannot fully erase the attack efficacy. This may be attributed to the deceptive command adversarially optimized on constructed defensive prompt.

\textbf{Visual Defense.} Compared to textual defense, the visual defense strategy shows limited effect. Applying Gaussian Blur achieves only a 2.8\% average reduction in ASR and is ineffective in two specific cases. This indicates that Gaussian Blur, while capable of suppressing some high-frequency adversarial signals, lacks the semantic grounding necessary to counteract cross-modal prompt injection. Unlike textual defenses that directly reinforce the agent's objective, visual defenses provide no explicit task guidance, leaving the agent vulnerable to semantically aligned visual cues.

\textbf{Combined Defense.} We further evaluate a combined defense strategy that integrates both sandwich prompting and Gaussian Blur. In two cases, the combined approach achieves better defensive performance than either method alone. However, the improvements are not additive, and in some scenarios, the combined effectiveness does not exceed that of the stronger individual defense. One possible explanation is that the two strategies may exhibit partially overlapping effects, introducing redundancy that limits the defense.

\begin{table}[!h]
    \centering
    \renewcommand\arraystretch{1.2}
    \scriptsize
    \resizebox{\linewidth}{!}{
        \begin{tabular}{c|c|c|c|c|c}
        \toprule
             Role & Model & Text Defense & Vision Defense & Combined Defense & No Defense  \\
              \midrule
              & Qwen2-VL & {82.0} & 95.0 & 85.7 & 97.0 \\
             \multirow{-2}{*}{RM} & Phi-3.5-vision  & 86.0 & 90.3 & {83.7} & 90.0 \\
              \midrule
              & Qwen2-VL & {83.0} & 84.0 & 85.0 & 84.0 \\
              \multirow{-2}{*}{PG} & Phi-3.5-vision & 83.3 & 80.7 & {71.0} & 90.0 \\
            \bottomrule[1.0pt]  
        \end{tabular}
    }
    \caption{Defense results (ASR \%) against our attack.}
    \label{tab:countermeasure}
\end{table}

%% file: 8_conclu.tex
\section{Conclusion and Future Work}
In this paper, we identify a critical yet previously overlooked security vulnerability in multimodal agents: cross-modal prompt injection attack, and propose \tool \ attack framework. Extensive experiments demonstrate that our method outperforms state-of-the-art attacks, achieving at least +26.4\% increase in attack success rates across diverse tasks. Additionally, we validate our attack's effectiveness in real-world multimodal autonomous agents. We hope that our work will inspire further investigations into the security of multimodal agents.

\textbf{Limitations}. Though promising, we would like to explore the following aspects in the future: \ding{182} designing more effective defense strategies such as adversarial training \cite{liu2023towards,zhang2021interpreting,liu2021training} for multimodal setting in open environments, and \ding{183} extending attack applicability to real-world agents with more complex architecture.

%% file: 0-main.bbl

\begin{thebibliography}{98}
\ifx \bisbn   \undefined \def \bisbn  #1{ISBN #1}\fi
\ifx \binits  \undefined \def \binits#1{#1}\fi
\ifx \bauthor  \undefined \def \bauthor#1{#1}\fi
\ifx \batitle  \undefined \def \batitle#1{#1}\fi
\ifx \bjtitle  \undefined \def \bjtitle#1{#1}\fi
\ifx \bvolume  \undefined \def \bvolume#1{\textbf{#1}}\fi
\ifx \byear  \undefined \def \byear#1{#1}\fi
\ifx \bissue  \undefined \def \bissue#1{#1}\fi
\ifx \bfpage  \undefined \def \bfpage#1{#1}\fi
\ifx \blpage  \undefined \def \blpage #1{#1}\fi
\ifx \burl  \undefined \def \burl#1{\textsf{#1}}\fi
\ifx \doiurl  \undefined \def \doiurl#1{\url{https://doi.org/#1}}\fi
\ifx \betal  \undefined \def \betal{\textit{et al.}}\fi
\ifx \binstitute  \undefined \def \binstitute#1{#1}\fi
\ifx \binstitutionaled  \undefined \def \binstitutionaled#1{#1}\fi
\ifx \bctitle  \undefined \def \bctitle#1{#1}\fi
\ifx \beditor  \undefined \def \beditor#1{#1}\fi
\ifx \bpublisher  \undefined \def \bpublisher#1{#1}\fi
\ifx \bbtitle  \undefined \def \bbtitle#1{#1}\fi
\ifx \bedition  \undefined \def \bedition#1{#1}\fi
\ifx \bseriesno  \undefined \def \bseriesno#1{#1}\fi
\ifx \blocation  \undefined \def \blocation#1{#1}\fi
\ifx \bsertitle  \undefined \def \bsertitle#1{#1}\fi
\ifx \bsnm \undefined \def \bsnm#1{#1}\fi
\ifx \bsuffix \undefined \def \bsuffix#1{#1}\fi
\ifx \bparticle \undefined \def \bparticle#1{#1}\fi
\ifx \barticle \undefined \def \barticle#1{#1}\fi
\bibcommenthead
\bibcommenthead
\bibcommenthead
\ifx \bconfdate \undefined \def \bconfdate #1{#1}\fi
\ifx \botherref \undefined \def \botherref #1{#1}\fi
\ifx \url \undefined \def \url#1{\textsf{#1}}\fi
\ifx \bchapter \undefined \def \bchapter#1{#1}\fi
\ifx \bbook \undefined \def \bbook#1{#1}\fi
\ifx \bcomment \undefined \def \bcomment#1{#1}\fi
\ifx \oauthor \undefined \def \oauthor#1{#1}\fi
\ifx \citeauthoryear \undefined \def \citeauthoryear#1{#1}\fi
\ifx \endbibitem  \undefined \def \endbibitem {}\fi
\ifx \bconflocation  \undefined \def \bconflocation#1{#1}\fi
\ifx \arxivurl  \undefined \def \arxivurl#1{\textsf{#1}}\fi
\csname PreBibitemsHook\endcsname

\bibitem[\protect\citeauthoryear{Xie et~al.}{2024}]{XieLMA2024}
\begin{botherref}
\oauthor{\bsnm{Xie}, \binits{J.}},
\oauthor{\bsnm{Chen}, \binits{Z.}},
\oauthor{\bsnm{Zhang}, \binits{R.}},
\oauthor{\bsnm{Wan}, \binits{X.}},
\oauthor{\bsnm{Li}, \binits{G.}}:
{Large Multimodal Agents: A Survey}.
arXiv preprint arXiv:2402.15116
(2024)
\end{botherref}
\endbibitem

\bibitem[\protect\citeauthoryear{Liu et~al.}{2023}]{LiuVisual2023}
\begin{bchapter}
\bauthor{\bsnm{Liu}, \binits{H.}},
\bauthor{\bsnm{Li}, \binits{C.}},
\bauthor{\bsnm{Wu}, \binits{Q.}},
\bauthor{\bsnm{Lee}, \binits{Y.J.}}:
\bctitle{{Visual Instruction Tuning}}.
In: \bbtitle{Advances in Neural Information Processing Systems}
(\byear{2023})
\end{bchapter}
\endbibitem

\bibitem[\protect\citeauthoryear{Liu et~al.}{2024}]{liuimproved2024}
\begin{bchapter}
\bauthor{\bsnm{Liu}, \binits{H.}},
\bauthor{\bsnm{Li}, \binits{C.}},
\bauthor{\bsnm{Li}, \binits{Y.}},
\bauthor{\bsnm{Lee}, \binits{Y.J.}}:
\bctitle{{Improved Baselines with Visual Instruction Tuning}}.
In: \bbtitle{Proceedings of the IEEE/CVF Conference on Computer Vision and Pattern Recognition}
(\byear{2024})
\end{bchapter}
\endbibitem

\bibitem[\protect\citeauthoryear{Dai et~al.}{2023}]{DaiInstructBlip2023}
\begin{bchapter}
\bauthor{\bsnm{Dai}, \binits{W.}},
\bauthor{\bsnm{Li}, \binits{J.}},
\bauthor{\bsnm{Li}, \binits{D.}},
\bauthor{\bsnm{Tiong}, \binits{A.}},
\bauthor{\bsnm{Zhao}, \binits{J.}},
\bauthor{\bsnm{Wang}, \binits{W.}},
\bauthor{\bsnm{Li}, \binits{B.}},
\bauthor{\bsnm{Fung}, \binits{P.}},
\bauthor{\bsnm{Hoi}, \binits{S.}}:
\bctitle{{Instruct{BLIP}: Towards General-purpose Vision-Language Models with Instruction Tuning}}.
In: \bbtitle{Advances in Neural Information Processing Systems}
(\byear{2023})
\end{bchapter}
\endbibitem

\bibitem[\protect\citeauthoryear{Zhuang et~al.}{2025}]{zhuangpuma2025}
\begin{bchapter}
\bauthor{\bsnm{Zhuang}, \binits{W.}},
\bauthor{\bsnm{Huang}, \binits{X.}},
\bauthor{\bsnm{Zhang}, \binits{X.}},
\bauthor{\bsnm{Zeng}, \binits{J.}}:
\bctitle{{Math-PUMA: Progressive Upward Multimodal Alignment to Enhance Mathematical Reasoning}}.
In: \bbtitle{Proceedings of the AAAI Conference on Artificial Intelligence}
(\byear{2025})
\end{bchapter}
\endbibitem

\bibitem[\protect\citeauthoryear{Chen et~al.}{2024}]{ChenInternVL2024}
\begin{bchapter}
\bauthor{\bsnm{Chen}, \binits{Z.}},
\bauthor{\bsnm{Wu}, \binits{J.}},
\bauthor{\bsnm{Wang}, \binits{W.}},
\bauthor{\bsnm{Su}, \binits{W.}},
\bauthor{\bsnm{Chen}, \binits{G.}},
\bauthor{\bsnm{Xing}, \binits{S.}},
\bauthor{\bsnm{Zhong}, \binits{M.}},
\bauthor{\bsnm{Zhang}, \binits{Q.}},
\bauthor{\bsnm{Zhu}, \binits{X.}},
\bauthor{\bsnm{Lu}, \binits{L.}},
\bauthor{\bsnm{Li}, \binits{B.}},
\bauthor{\bsnm{Luo}, \binits{P.}},
\bauthor{\bsnm{Lu}, \binits{T.}},
\bauthor{\bsnm{Qiao}, \binits{Y.}},
\bauthor{\bsnm{Dai}, \binits{J.}}:
\bctitle{{InternVL: Scaling up Vision Foundation Models and Aligning for Generic Visual-Linguistic Tasks}}.
In: \bbtitle{Proceedings of the IEEE/CVF Conference on Computer Vision and Pattern Recognition}
(\byear{2024})
\end{bchapter}
\endbibitem

\bibitem[\protect\citeauthoryear{Yao et~al.}{2024}]{YaoMinicpm2024}
\begin{botherref}
\oauthor{\bsnm{Yao}, \binits{Y.}},
\oauthor{\bsnm{Yu}, \binits{T.}},
\oauthor{\bsnm{Zhang}, \binits{A.}},
\oauthor{\bsnm{Wang}, \binits{C.}},
\oauthor{\bsnm{Cui}, \binits{J.}},
\oauthor{\bsnm{Zhu}, \binits{H.}},
\oauthor{\bsnm{Cai}, \binits{T.}},
\oauthor{\bsnm{Li}, \binits{H.}},
\oauthor{\bsnm{Zhao}, \binits{W.}},
\oauthor{\bsnm{He}, \binits{Z.}},
\oauthor{\bsnm{Chen}, \binits{Q.}},
\oauthor{\bsnm{Zhou}, \binits{H.}},
\oauthor{\bsnm{Zou}, \binits{Z.}},
\oauthor{\bsnm{Zhang}, \binits{H.}},
\oauthor{\bsnm{Hu}, \binits{S.}},
\oauthor{\bsnm{Zheng}, \binits{Z.}},
\oauthor{\bsnm{Zhou}, \binits{J.}},
\oauthor{\bsnm{Cai}, \binits{J.}},
\oauthor{\bsnm{Han}, \binits{X.}},
\oauthor{\bsnm{Zeng}, \binits{G.}},
\oauthor{\bsnm{Li}, \binits{D.}},
\oauthor{\bsnm{Liu}, \binits{Z.}},
\oauthor{\bsnm{Sun}, \binits{M.}}:
{MiniCPM-V: A GPT-4V Level MLLM on Your Phone}.
arXiv preprint arXiv:2408.01800
(2024)
\end{botherref}
\endbibitem

\bibitem[\protect\citeauthoryear{Wang et~al.}{2024}]{WangCogVLM2024}
\begin{botherref}
\oauthor{\bsnm{Wang}, \binits{W.}},
\oauthor{\bsnm{Lv}, \binits{Q.}},
\oauthor{\bsnm{Yu}, \binits{W.}},
\oauthor{\bsnm{Hong}, \binits{W.}},
\oauthor{\bsnm{Qi}, \binits{J.}},
\oauthor{\bsnm{Wang}, \binits{Y.}},
\oauthor{\bsnm{Ji}, \binits{J.}},
\oauthor{\bsnm{Yang}, \binits{Z.}},
\oauthor{\bsnm{Zhao}, \binits{L.}},
\oauthor{\bsnm{Song}, \binits{X.}},
\oauthor{\bsnm{Xu}, \binits{J.}},
\oauthor{\bsnm{Xu}, \binits{B.}},
\oauthor{\bsnm{Li}, \binits{J.}},
\oauthor{\bsnm{Dong}, \binits{Y.}},
\oauthor{\bsnm{Ding}, \binits{M.}},
\oauthor{\bsnm{Tang}, \binits{J.}}:
{CogVLM: Visual Expert for Pretrained Language Models}.
arXiv preprint arXiv:2311.03079
(2024)
\end{botherref}
\endbibitem

\bibitem[\protect\citeauthoryear{Pei et~al.}{2024}]{PeiWorkflow2024}
\begin{botherref}
\oauthor{\bsnm{Pei}, \binits{J.}},
\oauthor{\bsnm{Viola}, \binits{I.}},
\oauthor{\bsnm{Huang}, \binits{H.}},
\oauthor{\bsnm{Wang}, \binits{J.}},
\oauthor{\bsnm{Ahsan}, \binits{M.}},
\oauthor{\bsnm{Ye}, \binits{F.}},
\oauthor{\bsnm{Yiming}, \binits{J.}},
\oauthor{\bsnm{Sai}, \binits{Y.}},
\oauthor{\bsnm{Wang}, \binits{D.}},
\oauthor{\bsnm{Chen}, \binits{Z.}},
\oauthor{\bsnm{Ren}, \binits{P.}},
\oauthor{\bsnm{Cesar}, \binits{P.}}:
{Autonomous Workflow for Multimodal Fine-Grained Training Assistants Towards Mixed Reality}.
arXiv preprint arXiv:2405.13034
(2024)
\end{botherref}
\endbibitem

\bibitem[\protect\citeauthoryear{Chu et~al.}{2023}]{ChuMobile2023}
\begin{botherref}
\oauthor{\bsnm{Chu}, \binits{X.}},
\oauthor{\bsnm{Qiao}, \binits{L.}},
\oauthor{\bsnm{Lin}, \binits{X.}},
\oauthor{\bsnm{Xu}, \binits{S.}},
\oauthor{\bsnm{Yang}, \binits{Y.}},
\oauthor{\bsnm{Hu}, \binits{Y.}},
\oauthor{\bsnm{Wei}, \binits{F.}},
\oauthor{\bsnm{Zhang}, \binits{X.}},
\oauthor{\bsnm{Zhang}, \binits{B.}},
\oauthor{\bsnm{Wei}, \binits{X.}},
\oauthor{\bsnm{Shen}, \binits{C.}}:
{MobileVLM: A Fast, Strong and Open Vision Language Assistant for Mobile Devices}.
arXiv preprint arXiv:2312.16886
(2023)
\end{botherref}
\endbibitem

\bibitem[\protect\citeauthoryear{Hong et~al.}{2024}]{HongCogAgent2024}
\begin{bchapter}
\bauthor{\bsnm{Hong}, \binits{W.}},
\bauthor{\bsnm{Wang}, \binits{W.}},
\bauthor{\bsnm{Lv}, \binits{Q.}},
\bauthor{\bsnm{Xu}, \binits{J.}},
\bauthor{\bsnm{Yu}, \binits{W.}},
\bauthor{\bsnm{Ji}, \binits{J.}},
\bauthor{\bsnm{Wang}, \binits{Y.}},
\bauthor{\bsnm{Wang}, \binits{Z.}},
\bauthor{\bsnm{Dong}, \binits{Y.}},
\bauthor{\bsnm{Ding}, \binits{M.}},
\bauthor{\bsnm{Tang}, \binits{J.}}:
\bctitle{{CogAgent: A Visual Language Model for GUI Agents}}.
In: \bbtitle{Proceedings of the IEEE/CVF Conference on Computer Vision and Pattern Recognition}
(\byear{2024})
\end{bchapter}
\endbibitem

\bibitem[\protect\citeauthoryear{Wang et~al.}{2023}]{WangDriveMLM2023}
\begin{botherref}
\oauthor{\bsnm{Wang}, \binits{W.}},
\oauthor{\bsnm{Xie}, \binits{J.}},
\oauthor{\bsnm{Hu}, \binits{C.}},
\oauthor{\bsnm{Zou}, \binits{H.}},
\oauthor{\bsnm{Fan}, \binits{J.}},
\oauthor{\bsnm{Tong}, \binits{W.}},
\oauthor{\bsnm{Wen}, \binits{Y.}},
\oauthor{\bsnm{Wu}, \binits{S.}},
\oauthor{\bsnm{Deng}, \binits{H.}},
\oauthor{\bsnm{Li}, \binits{Z.}},
\oauthor{\bsnm{Tian}, \binits{H.}},
\oauthor{\bsnm{Lu}, \binits{L.}},
\oauthor{\bsnm{Zhu}, \binits{X.}},
\oauthor{\bsnm{Wang}, \binits{X.}},
\oauthor{\bsnm{Qiao}, \binits{Y.}},
\oauthor{\bsnm{Dai}, \binits{J.}}:
{DriveMLM: Aligning Multi-Modal Large Language Models with Behavioral Planning States for Autonomous Driving}.
arXiv preprint arXiv:2312.09245
(2023)
\end{botherref}
\endbibitem

\bibitem[\protect\citeauthoryear{Ma et~al.}{2024}]{MaDolphin2024}
\begin{bchapter}
\bauthor{\bsnm{Ma}, \binits{Y.}},
\bauthor{\bsnm{Cao}, \binits{Y.}},
\bauthor{\bsnm{Sun}, \binits{J.}},
\bauthor{\bsnm{Pavone}, \binits{M.}},
\bauthor{\bsnm{Xiao}, \binits{C.}}:
\bctitle{Dolphins: Multimodal language model for driving}.
In: \bbtitle{European Conference on Computer Vision}
(\byear{2024})
\end{bchapter}
\endbibitem

\bibitem[\protect\citeauthoryear{Xu et~al.}{2024}]{XuVLMAD2024}
\begin{botherref}
\oauthor{\bsnm{Xu}, \binits{Y.}},
\oauthor{\bsnm{Hu}, \binits{Y.}},
\oauthor{\bsnm{Zhang}, \binits{Z.}},
\oauthor{\bsnm{Meyer}, \binits{G.P.}},
\oauthor{\bsnm{Mustikovela}, \binits{S.K.}},
\oauthor{\bsnm{Srinivasa}, \binits{S.}},
\oauthor{\bsnm{Wolff}, \binits{E.M.}},
\oauthor{\bsnm{Huang}, \binits{X.}}:
{VLM-AD: End-to-End Autonomous Driving through Vision-Language Model Supervision}.
arXiv preprint arXiv:2412.14446
(2024)
\end{botherref}
\endbibitem

\bibitem[\protect\citeauthoryear{Abdin et~al.}{2024}]{Phi}
\begin{botherref}
\oauthor{\bsnm{Abdin}, \binits{M.}},
\oauthor{\bsnm{Aneja}, \binits{J.}},
\oauthor{\bsnm{Hany~Awadalla}, \binits{e.a.}}:
{Phi-3 Technical Report: A Highly Capable Language Model Locally on Your Phone}.
arXiv preprint arXiv:2404.14219
(2024)
\end{botherref}
\endbibitem

\bibitem[\protect\citeauthoryear{Wallace et~al.}{2024}]{KimOpenVLA2024}
\begin{botherref}
\oauthor{\bsnm{Wallace}, \binits{E.}},
\oauthor{\bsnm{Xiao}, \binits{K.}},
\oauthor{\bsnm{Leike}, \binits{R.}},
\oauthor{\bsnm{Weng}, \binits{L.}},
\oauthor{\bsnm{Heidecke}, \binits{J.}},
\oauthor{\bsnm{Beutel}, \binits{A.}}:
{OpenVLA: An Open-Source Vision-Language-Action Model}.
arXiv preprint arXiv:2404.13208
(2024)
\end{botherref}
\endbibitem

\bibitem[\protect\citeauthoryear{Mu et~al.}{2023}]{MuEmbodied2023}
\begin{bchapter}
\bauthor{\bsnm{Mu}, \binits{Y.}},
\bauthor{\bsnm{Zhang}, \binits{Q.}},
\bauthor{\bsnm{Hu}, \binits{M.}},
\bauthor{\bsnm{Wang}, \binits{W.}},
\bauthor{\bsnm{Ding}, \binits{M.}},
\bauthor{\bsnm{Jin}, \binits{J.}},
\bauthor{\bsnm{Wang}, \binits{B.}},
\bauthor{\bsnm{Dai}, \binits{J.}},
\bauthor{\bsnm{Qiao}, \binits{Y.}},
\bauthor{\bsnm{Luo}, \binits{P.}}:
\bctitle{Embodiedgpt: Vision-language pre-training via embodied chain of thought}.
In: \bbtitle{Advances in Neural Information Processing Systems}
(\byear{2023})
\end{bchapter}
\endbibitem

\bibitem[\protect\citeauthoryear{Li et~al.}{2024}]{LiVLA2024}
\begin{bchapter}
\bauthor{\bsnm{Li}, \binits{X.}},
\bauthor{\bsnm{Liu}, \binits{M.}},
\bauthor{\bsnm{Zhang}, \binits{H.}},
\bauthor{\bsnm{Yu}, \binits{C.}},
\bauthor{\bsnm{Xu}, \binits{J.}},
\bauthor{\bsnm{Wu}, \binits{H.}},
\bauthor{\bsnm{Cheang}, \binits{C.}},
\bauthor{\bsnm{Jing}, \binits{Y.}},
\bauthor{\bsnm{Zhang}, \binits{W.}},
\bauthor{\bsnm{Liu}, \binits{H.}},
\bauthor{\bsnm{Li}, \binits{H.}},
\bauthor{\bsnm{Kong}, \binits{T.}}:
\bctitle{Vision-language foundation models as effective robot imitators}.
In: \bbtitle{International Conference on Learning Representations}
(\byear{2024})
\end{bchapter}
\endbibitem

\bibitem[\protect\citeauthoryear{Qin et~al.}{2023}]{qinmp52024}
\begin{botherref}
\oauthor{\bsnm{Qin}, \binits{Y.}},
\oauthor{\bsnm{Zhou}, \binits{E.}},
\oauthor{\bsnm{Liu}, \binits{Q.}},
\oauthor{\bsnm{Yin}, \binits{Z.}},
\oauthor{\bsnm{Sheng}, \binits{L.}},
\oauthor{\bsnm{Zhang}, \binits{R.}},
\oauthor{\bsnm{Qiao}, \binits{Y.}},
\oauthor{\bsnm{Shao}, \binits{J.}}:
{MP5: A Multi-modal Open-ended Embodied System in Minecraft via Active Perception}.
arXiv preprint arXiv:2312.07472
(2023)
\end{botherref}
\endbibitem

\bibitem[\protect\citeauthoryear{{OWASP}}{2023}]{OWASP}
\begin{botherref}
\oauthor{\bsnm{{OWASP}}}:
{OWASP Top 10 for LLM Applications}.
\url{https://owasp.org/www-project-top-10-for-large-language-model-applications/assets/PDF/OWASP-Top-10-for-LLMs-2023-v1_1.pdf}
\end{botherref}
\endbibitem

\bibitem[\protect\citeauthoryear{Shayegani et~al.}{2024}]{ErfanJIP2024}
\begin{bchapter}
\bauthor{\bsnm{Shayegani}, \binits{E.}},
\bauthor{\bsnm{Dong}, \binits{Y.}},
\bauthor{\bsnm{Abu{-}Ghazaleh}, \binits{N.B.}}:
\bctitle{{J}ailbreak in pieces: Compositional adversarial attacks on multi-modal language models}.
In: \bbtitle{International Conference on Learning Representations}
(\byear{2024})
\end{bchapter}
\endbibitem

\bibitem[\protect\citeauthoryear{Wu et~al.}{2024}]{WuAgent2024}
\begin{botherref}
\oauthor{\bsnm{Wu}, \binits{C.H.}},
\oauthor{\bsnm{Shah}, \binits{R.}},
\oauthor{\bsnm{Koh}, \binits{J.Y.}},
\oauthor{\bsnm{Salakhutdinov}, \binits{R.}},
\oauthor{\bsnm{Fried}, \binits{D.}},
\oauthor{\bsnm{Raghunathan}, \binits{A.}}:
{Dissecting Adversarial Robustness of Multimodal LM Agents}.
arXiv preprint arXiv:2406.12814
(2024)
\end{botherref}
\endbibitem

\bibitem[\protect\citeauthoryear{Liu et~al.}{2024}]{LiuAuto2024}
\begin{botherref}
\oauthor{\bsnm{Liu}, \binits{X.}},
\oauthor{\bsnm{Yu}, \binits{Z.}},
\oauthor{\bsnm{Zhang}, \binits{Y.}},
\oauthor{\bsnm{Zhang}, \binits{N.}},
\oauthor{\bsnm{Xiao}, \binits{C.}}:
{Automatic and Universal Prompt Injection Attacks against Large Language Modals}.
arXiv preprint arXiv:2403.04957
(2024)
\end{botherref}
\endbibitem

\bibitem[\protect\citeauthoryear{Liu et~al.}{2023}]{LiuPrompt2023}
\begin{botherref}
\oauthor{\bsnm{Liu}, \binits{Y.}},
\oauthor{\bsnm{Deng}, \binits{G.}},
\oauthor{\bsnm{Li}, \binits{Y.}},
\oauthor{\bsnm{Wang}, \binits{K.}},
\oauthor{\bsnm{Zhang}, \binits{T.}},
\oauthor{\bsnm{Liu}, \binits{Y.}},
\oauthor{\bsnm{Wang}, \binits{H.}},
\oauthor{\bsnm{Zheng}, \binits{Y.}},
\oauthor{\bsnm{Liu}, \binits{Y.}}:
{Prompt Injection attack against LLM-integrated Applications}.
arXiv preprint arXiv:2306.05499
(2023)
\end{botherref}
\endbibitem

\bibitem[\protect\citeauthoryear{Abdelnabi et~al.}{2023}]{SaharCompromise2024}
\begin{bchapter}
\bauthor{\bsnm{Abdelnabi}, \binits{S.}},
\bauthor{\bsnm{Greshake}, \binits{K.}},
\bauthor{\bsnm{Mishra}, \binits{S.}},
\bauthor{\bsnm{Endres}, \binits{C.}},
\bauthor{\bsnm{Holz}, \binits{T.}},
\bauthor{\bsnm{Fritz}, \binits{M.}}:
\bctitle{Not what you've signed up for: Compromising real-world llm-integrated applications with indirect prompt injection}.
In: \bbtitle{{ACM} Workshop on Artificial Intelligence and Security, AISec}
(\byear{2023})
\end{bchapter}
\endbibitem

\bibitem[\protect\citeauthoryear{Yi et~al.}{2025}]{YiIndirect2025}
\begin{botherref}
\oauthor{\bsnm{Yi}, \binits{J.}},
\oauthor{\bsnm{Xie}, \binits{Y.}},
\oauthor{\bsnm{Zhu}, \binits{B.}},
\oauthor{\bsnm{Kiciman}, \binits{E.}},
\oauthor{\bsnm{Sun}, \binits{G.}},
\oauthor{\bsnm{Xie}, \binits{X.}},
\oauthor{\bsnm{Wu}, \binits{F.}}:
{Benchmarking and Defending Against Indirect Prompt Injection Attacks on Large Language Models}.
arXiv preprint arXiv:2312.14197
(2025)
\end{botherref}
\endbibitem

\bibitem[\protect\citeauthoryear{Zhan et~al.}{2024}]{ZhanInjec2024}
\begin{botherref}
\oauthor{\bsnm{Zhan}, \binits{Q.}},
\oauthor{\bsnm{Liang}, \binits{Z.}},
\oauthor{\bsnm{Ying}, \binits{Z.}},
\oauthor{\bsnm{Kang}, \binits{D.}}:
Injecagent: Benchmarking indirect prompt injections in tool-integrated large language model agents.
arXiv preprint arXiv:2403.02691
(2024)
\end{botherref}
\endbibitem

\bibitem[\protect\citeauthoryear{Sur{\'\i}s et~al.}{2023}]{SurisViper2023}
\begin{bchapter}
\bauthor{\bsnm{Sur{\'\i}s}, \binits{D.}},
\bauthor{\bsnm{Menon}, \binits{S.}},
\bauthor{\bsnm{Vondrick}, \binits{C.}}:
\bctitle{{ViperGPT: Visual Inference via Python Execution for Reasoning}}.
In: \bbtitle{Proceedings of the IEEE/CVF International Conference on Computer Vision}
(\byear{2023})
\end{bchapter}
\endbibitem

\bibitem[\protect\citeauthoryear{Pan et~al.}{2023}]{LuChameleon2023}
\begin{bchapter}
\bauthor{\bsnm{Pan}, \binits{L.}},
\bauthor{\bsnm{Baolin}, \binits{P.}},
\bauthor{\bsnm{Hao}, \binits{C.}},
\bauthor{\bsnm{Michel}, \binits{G.}},
\bauthor{\bsnm{Kai-Wei}, \binits{C.}},
\bauthor{\bsnm{Wu}, \binits{Y.N.}},
\bauthor{\bsnm{Zhu1}, \binits{S.-C.}},
\bauthor{\bsnm{Gao2}, \binits{J.}}:
\bctitle{{Chameleon: Plug-and-Play Compositional Reasoning with Large Language Models}}.
In: \bbtitle{Advances in Neural Information Processing Systems}
(\byear{2023})
\end{bchapter}
\endbibitem

\bibitem[\protect\citeauthoryear{Gupta and Kembhavi}{2023}]{GuptaVisProg2023}
\begin{bchapter}
\bauthor{\bsnm{Gupta}, \binits{T.}},
\bauthor{\bsnm{Kembhavi}, \binits{A.}}:
\bctitle{{Visual Programming: Compositional Visual Reasoning Without Training}}.
In: \bbtitle{Proceedings of the IEEE/CVF Conference on Computer Vision and Pattern Recognition}
(\byear{2023})
\end{bchapter}
\endbibitem

\bibitem[\protect\citeauthoryear{Wu et~al.}{2023}]{WuVisualGPT2023}
\begin{botherref}
\oauthor{\bsnm{Wu}, \binits{C.}},
\oauthor{\bsnm{Yin}, \binits{S.}},
\oauthor{\bsnm{Qi}, \binits{W.}},
\oauthor{\bsnm{Wang}, \binits{X.}},
\oauthor{\bsnm{Tang}, \binits{Z.}},
\oauthor{\bsnm{Duan}, \binits{N.}}:
{Visual ChatGPT: Talking, Drawing and Editing with Visual Foundation Models}.
arXiv preprint arXiv:2303.04671
(2023)
\end{botherref}
\endbibitem

\bibitem[\protect\citeauthoryear{Black et~al.}{2024}]{KevinPi02024}
\begin{botherref}
\oauthor{\bsnm{Black}, \binits{K.}},
\oauthor{\bsnm{Brown}, \binits{N.}},
\oauthor{\bsnm{Driess}, \binits{D.}},
\oauthor{\bsnm{Esmail}, \binits{A.}},
\oauthor{\bsnm{Equi}, \binits{M.}},
\oauthor{\bsnm{Finn}, \binits{C.}},
\oauthor{\bsnm{Fusai}, \binits{N.}},
\oauthor{\bsnm{Groom}, \binits{L.}},
\oauthor{\bsnm{Hausman}, \binits{K.}},
\oauthor{\bsnm{Ichter}, \binits{B.}},
\oauthor{\bsnm{Jakubczak}, \binits{S.}},
\oauthor{\bsnm{Jones}, \binits{T.}},
\oauthor{\bsnm{Ke}, \binits{L.}},
\oauthor{\bsnm{Levine}, \binits{S.}},
\oauthor{\bsnm{Li-Bell}, \binits{A.}},
\oauthor{\bsnm{Mothukuri}, \binits{M.}},
\oauthor{\bsnm{Nair}, \binits{S.}},
\oauthor{\bsnm{Pertsch}, \binits{K.}},
\oauthor{\bsnm{Shi}, \binits{L.X.}},
\oauthor{\bsnm{Tanner}, \binits{J.}},
\oauthor{\bsnm{Vuong}, \binits{Q.}},
\oauthor{\bsnm{Walling}, \binits{A.}},
\oauthor{\bsnm{Wang}, \binits{H.}},
\oauthor{\bsnm{Zhilinsky}, \binits{U.}}:
{$\pi_0$: A Vision-Language-Action Flow Model for General Robot Control}.
arXiv preprint arXiv:2410.24164
(2024)
\end{botherref}
\endbibitem

\bibitem[\protect\citeauthoryear{Lan et~al.}{2025}]{LanBFA2025}
\begin{botherref}
\oauthor{\bsnm{Lan}, \binits{Z.}},
\oauthor{\bsnm{Mao}, \binits{W.}},
\oauthor{\bsnm{Li}, \binits{H.}},
\oauthor{\bsnm{Wang}, \binits{L.}},
\oauthor{\bsnm{Wang}, \binits{T.}},
\oauthor{\bsnm{Fan}, \binits{H.}},
\oauthor{\bsnm{Yoshie}, \binits{O.}}:
{BFA: Best-Feature-Aware Fusion for Multi-View Fine-grained Manipulation}.
arXiv preprint arxiv:2502.11161
(2025)
\end{botherref}
\endbibitem

\bibitem[\protect\citeauthoryear{Touvron et~al.}{2023}]{touvronLlama2023}
\begin{botherref}
\oauthor{\bsnm{Touvron}, \binits{H.}},
\oauthor{\bsnm{Lavril}, \binits{T.}},
\oauthor{\bsnm{Izacard}, \binits{G.}},
\oauthor{\bsnm{Martinet}, \binits{X.}},
\oauthor{\bsnm{Lachaux}, \binits{M.-A.}},
\oauthor{\bsnm{Lacroix}, \binits{T.}},
\oauthor{\bsnm{Rozière}, \binits{B.}},
\oauthor{\bsnm{Goyal}, \binits{N.}},
\oauthor{\bsnm{Hambro}, \binits{E.}},
\oauthor{\bsnm{Azhar}, \binits{F.}},
\oauthor{\bsnm{Rodriguez}, \binits{A.}},
\oauthor{\bsnm{Joulin}, \binits{A.}},
\oauthor{\bsnm{Grave}, \binits{E.}},
\oauthor{\bsnm{Lample}, \binits{G.}}:
{LLaMA: Open and Efficient Foundation Language Models}.
arXiv preprint arXiv:2302.13971
(2023)
\end{botherref}
\endbibitem

\bibitem[\protect\citeauthoryear{Liu et~al.}{2024}]{LiuCompromising2024}
\begin{botherref}
\oauthor{\bsnm{Liu}, \binits{A.}},
\oauthor{\bsnm{Zhou}, \binits{Y.}},
\oauthor{\bsnm{Liu}, \binits{X.}},
\oauthor{\bsnm{Zhang}, \binits{T.}},
\oauthor{\bsnm{Liang}, \binits{S.}},
\oauthor{\bsnm{Wang}, \binits{J.}},
\oauthor{\bsnm{Pu}, \binits{Y.}},
\oauthor{\bsnm{Li}, \binits{T.}},
\oauthor{\bsnm{Zhang}, \binits{J.}},
\oauthor{\bsnm{Zhou}, \binits{W.}},
\oauthor{\bsnm{Guo}, \binits{Q.}},
\oauthor{\bsnm{Tao}, \binits{D.}}:
{Compromising Embodied Agents with Contextual Backdoor Attacks}.
arXiv preprint arXiv:2408.02882
(2024)
\end{botherref}
\endbibitem

\bibitem[\protect\citeauthoryear{Wang et~al.}{2025}]{WangTrojan2025}
\begin{botherref}
\oauthor{\bsnm{Wang}, \binits{X.}},
\oauthor{\bsnm{Pan}, \binits{H.}},
\oauthor{\bsnm{Zhang}, \binits{H.}},
\oauthor{\bsnm{Li}, \binits{M.}},
\oauthor{\bsnm{Hu}, \binits{S.}},
\oauthor{\bsnm{Zhou}, \binits{Z.}},
\oauthor{\bsnm{Xue}, \binits{L.}},
\oauthor{\bsnm{Guo}, \binits{P.}},
\oauthor{\bsnm{Wang}, \binits{Y.}},
\oauthor{\bsnm{Wan}, \binits{W.}},
\oauthor{\bsnm{Liu}, \binits{A.}},
\oauthor{\bsnm{Zhang}, \binits{L.Y.}}:
{TrojanRobot: Physical-World Backdoor Attacks Against VLM-based Robotic Manipulation}.
arXiv preprint arXiv:2411.11683
(2025)
\end{botherref}
\endbibitem

\bibitem[\protect\citeauthoryear{Zhang et~al.}{2025}]{ZhangBad2025}
\begin{botherref}
\oauthor{\bsnm{Zhang}, \binits{H.}},
\oauthor{\bsnm{Zhu}, \binits{C.}},
\oauthor{\bsnm{Wang}, \binits{X.}},
\oauthor{\bsnm{Zhou}, \binits{Z.}},
\oauthor{\bsnm{Yin}, \binits{C.}},
\oauthor{\bsnm{Li}, \binits{M.}},
\oauthor{\bsnm{Xue}, \binits{L.}},
\oauthor{\bsnm{Wang}, \binits{Y.}},
\oauthor{\bsnm{Hu}, \binits{S.}},
\oauthor{\bsnm{Liu}, \binits{A.}},
\oauthor{\bsnm{Guo}, \binits{P.}},
\oauthor{\bsnm{Zhang}, \binits{L.Y.}}:
{BadRobot: Jailbreaking Embodied LLMs in the Physical World}.
arXiv preprint arXiv:2407.20242
(2025)
\end{botherref}
\endbibitem

\bibitem[\protect\citeauthoryear{Aichberger et~al.}{2025}]{AichbergerOS2025}
\begin{botherref}
\oauthor{\bsnm{Aichberger}, \binits{L.}},
\oauthor{\bsnm{Paren}, \binits{A.}},
\oauthor{\bsnm{Gal}, \binits{Y.}},
\oauthor{\bsnm{Torr}, \binits{P.}},
\oauthor{\bsnm{Bibi}, \binits{A.}}:
{Attacking Multimodal OS Agents with Malicious Image Patches}.
arXiv preprint arXiv:2503.10809
(2025)
\end{botherref}
\endbibitem

\bibitem[\protect\citeauthoryear{Liu et~al.}{2020}]{liuspatiotemporal2020}
\begin{bchapter}
\bauthor{\bsnm{Liu}, \binits{A.}},
\bauthor{\bsnm{Huang}, \binits{T.}},
\bauthor{\bsnm{Liu}, \binits{X.}},
\bauthor{\bsnm{Xu}, \binits{Y.}},
\bauthor{\bsnm{Ma}, \binits{Y.}},
\bauthor{\bsnm{Chen}, \binits{X.}},
\bauthor{\bsnm{Maybank}, \binits{S.J.}},
\bauthor{\bsnm{Tao}, \binits{D.}}:
\bctitle{{Spatiotemporal attacks for embodied agents}}.
In: \bbtitle{European Conference on Computer Vision}
(\byear{2020})
\end{bchapter}
\endbibitem

\bibitem[\protect\citeauthoryear{Liang et~al.}{2020}]{liang2020efficient}
\begin{bchapter}
\bauthor{\bsnm{Liang}, \binits{S.}},
\bauthor{\bsnm{Wei}, \binits{X.}},
\bauthor{\bsnm{Yao}, \binits{S.}},
\bauthor{\bsnm{Cao}, \binits{X.}}:
\bctitle{{Efficient adversarial attacks for visual object tracking}}.
In: \bbtitle{Computer Vision--ECCV 2020: 16th European Conference, Glasgow, UK, August 23--28, 2020, Proceedings, Part XXVI 16}
(\byear{2020})
\end{bchapter}
\endbibitem

\bibitem[\protect\citeauthoryear{Wei et~al.}{2018}]{wei2018transferable}
\begin{botherref}
\oauthor{\bsnm{Wei}, \binits{X.}},
\oauthor{\bsnm{Liang}, \binits{S.}},
\oauthor{\bsnm{Chen}, \binits{N.}},
\oauthor{\bsnm{Cao}, \binits{X.}}:
{Transferable adversarial attacks for image and video object detection}.
arXiv preprint arXiv:1811.12641
(2018)
\end{botherref}
\endbibitem

\bibitem[\protect\citeauthoryear{Liang et~al.}{2022a}]{liang2022parallel}
\begin{botherref}
\oauthor{\bsnm{Liang}, \binits{S.}},
\oauthor{\bsnm{Wu}, \binits{B.}},
\oauthor{\bsnm{Fan}, \binits{Y.}},
\oauthor{\bsnm{Wei}, \binits{X.}},
\oauthor{\bsnm{Cao}, \binits{X.}}:
{Parallel rectangle flip attack: A query-based black-box attack against object detection}.
arXiv preprint arXiv:2201.08970
(2022)
\end{botherref}
\endbibitem

\bibitem[\protect\citeauthoryear{Liang et~al.}{2022b}]{liang2022large}
\begin{bchapter}
\bauthor{\bsnm{Liang}, \binits{S.}},
\bauthor{\bsnm{Li}, \binits{L.}},
\bauthor{\bsnm{Fan}, \binits{Y.}},
\bauthor{\bsnm{Jia}, \binits{X.}},
\bauthor{\bsnm{Li}, \binits{J.}},
\bauthor{\bsnm{Wu}, \binits{B.}},
\bauthor{\bsnm{Cao}, \binits{X.}}:
\bctitle{{A large-scale multiple-objective method for black-box attack against object detection}}.
In: \bbtitle{European Conference on Computer Vision}
(\byear{2022})
\end{bchapter}
\endbibitem

\bibitem[\protect\citeauthoryear{Yan et~al.}{2024}]{yan2024df40}
\begin{botherref}
\oauthor{\bsnm{Yan}, \binits{Z.}},
\oauthor{\bsnm{Yao}, \binits{T.}},
\oauthor{\bsnm{Chen}, \binits{S.}},
\oauthor{\bsnm{Zhao}, \binits{Y.}},
\oauthor{\bsnm{Fu}, \binits{X.}},
\oauthor{\bsnm{Zhu}, \binits{J.}},
\oauthor{\bsnm{Luo}, \binits{D.}},
\oauthor{\bsnm{Wang}, \binits{C.}},
\oauthor{\bsnm{Ding}, \binits{S.}},
\oauthor{\bsnm{Wu}, \binits{Y.}}, et al.:
Df40: Toward next-generation deepfake detection.
arXiv preprint arXiv:2406.13495
(2024)
\end{botherref}
\endbibitem

\bibitem[\protect\citeauthoryear{Ying and Wu}{2023a}]{Ying_2023}
\begin{botherref}
\oauthor{\bsnm{Ying}, \binits{Z.}},
\oauthor{\bsnm{Wu}, \binits{B.}}:
{DLP: towards active defense against backdoor attacks with decoupled learning process}.
Cybersecurity
\textbf{6}(1)
(2023)
\doiurl{10.1186/s42400-023-00141-4}
\end{botherref}
\endbibitem

\bibitem[\protect\citeauthoryear{Ying and Wu}{2023b}]{Ying_20231}
\begin{botherref}
\oauthor{\bsnm{Ying}, \binits{Z.}},
\oauthor{\bsnm{Wu}, \binits{B.}}:
{NBA: defensive distillation for backdoor removal via neural behavior alignment}.
Cybersecurity
\textbf{6}(1)
(2023)
\doiurl{10.1186/s42400-023-00154-z}
\end{botherref}
\endbibitem

\bibitem[\protect\citeauthoryear{Zhang et~al.}{2024}]{zhang2024visual}
\begin{botherref}
\oauthor{\bsnm{Zhang}, \binits{T.}},
\oauthor{\bsnm{Wang}, \binits{L.}},
\oauthor{\bsnm{Zhang}, \binits{X.}},
\oauthor{\bsnm{Zhang}, \binits{Y.}},
\oauthor{\bsnm{Jia}, \binits{B.}},
\oauthor{\bsnm{Liang}, \binits{S.}},
\oauthor{\bsnm{Hu}, \binits{S.}},
\oauthor{\bsnm{Fu}, \binits{Q.}},
\oauthor{\bsnm{Liu}, \binits{A.}},
\oauthor{\bsnm{Liu}, \binits{X.}}:
{Visual Adversarial Attack on Vision-Language Models for Autonomous Driving}.
arXiv preprint arXiv:2411.18275
(2024)
\end{botherref}
\endbibitem

\bibitem[\protect\citeauthoryear{Kong et~al.}{2024}]{kong2024patch}
\begin{barticle}
\bauthor{\bsnm{Kong}, \binits{D.}},
\bauthor{\bsnm{Liang}, \binits{S.}},
\bauthor{\bsnm{Zhu}, \binits{X.}},
\bauthor{\bsnm{Zhong}, \binits{Y.}},
\bauthor{\bsnm{Ren}, \binits{W.}}:
\batitle{Patch is enough: naturalistic adversarial patch against vision-language pre-training models}.
\bjtitle{Visual Intelligence}
\bvolume{2}(\bissue{1}),
\bfpage{1}--\blpage{10}
(\byear{2024})
\end{barticle}
\endbibitem

\bibitem[\protect\citeauthoryear{Ying et~al.}{2025}]{ying2025reasoning}
\begin{botherref}
\oauthor{\bsnm{Ying}, \binits{Z.}},
\oauthor{\bsnm{Zhang}, \binits{D.}},
\oauthor{\bsnm{Jing}, \binits{Z.}},
\oauthor{\bsnm{Xiao}, \binits{Y.}},
\oauthor{\bsnm{Zou}, \binits{Q.}},
\oauthor{\bsnm{Liu}, \binits{A.}},
\oauthor{\bsnm{Liang}, \binits{S.}},
\oauthor{\bsnm{Zhang}, \binits{X.}},
\oauthor{\bsnm{Liu}, \binits{X.}},
\oauthor{\bsnm{Tao}, \binits{D.}}:
{Reasoning-augmented conversation for multi-turn jailbreak attacks on large language models}.
arXiv preprint arXiv:2502.11054
(2025)
\end{botherref}
\endbibitem

\bibitem[\protect\citeauthoryear{Liang et~al.}{2023}]{liang2023badclip}
\begin{botherref}
\oauthor{\bsnm{Liang}, \binits{S.}},
\oauthor{\bsnm{Zhu}, \binits{M.}},
\oauthor{\bsnm{Liu}, \binits{A.}},
\oauthor{\bsnm{Wu}, \binits{B.}},
\oauthor{\bsnm{Cao}, \binits{X.}},
\oauthor{\bsnm{Chang}, \binits{E.-C.}}:
{BadClip: Dual-embedding guided backdoor attack on multimodal contrastive learning}.
arXiv preprint arXiv:2311.12075
(2023)
\end{botherref}
\endbibitem

\bibitem[\protect\citeauthoryear{Liang et~al.}{2024a}]{liang2024revisiting}
\begin{botherref}
\oauthor{\bsnm{Liang}, \binits{S.}},
\oauthor{\bsnm{Liang}, \binits{J.}},
\oauthor{\bsnm{Pang}, \binits{T.}},
\oauthor{\bsnm{Du}, \binits{C.}},
\oauthor{\bsnm{Liu}, \binits{A.}},
\oauthor{\bsnm{Chang}, \binits{E.-C.}},
\oauthor{\bsnm{Cao}, \binits{X.}}:
{Revisiting Backdoor Attacks against Large Vision-Language Models}.
arXiv preprint arXiv:2406.18844
(2024)
\end{botherref}
\endbibitem

\bibitem[\protect\citeauthoryear{Liang et~al.}{2024b}]{liang2024vl}
\begin{botherref}
\oauthor{\bsnm{Liang}, \binits{J.}},
\oauthor{\bsnm{Liang}, \binits{S.}},
\oauthor{\bsnm{Luo}, \binits{M.}},
\oauthor{\bsnm{Liu}, \binits{A.}},
\oauthor{\bsnm{Han}, \binits{D.}},
\oauthor{\bsnm{Chang}, \binits{E.-C.}},
\oauthor{\bsnm{Cao}, \binits{X.}}:
{VL-Trojan: Multimodal Instruction Backdoor Attacks against Autoregressive Visual Language Models}.
arXiv preprint arXiv:2402.13851
(2024)
\end{botherref}
\endbibitem

\bibitem[\protect\citeauthoryear{Ying et~al.}{2024a}]{ying2024jailbreak}
\begin{botherref}
\oauthor{\bsnm{Ying}, \binits{Z.}},
\oauthor{\bsnm{Liu}, \binits{A.}},
\oauthor{\bsnm{Zhang}, \binits{T.}},
\oauthor{\bsnm{Yu}, \binits{Z.}},
\oauthor{\bsnm{Liang}, \binits{S.}},
\oauthor{\bsnm{Liu}, \binits{X.}},
\oauthor{\bsnm{Tao}, \binits{D.}}:
{Jailbreak vision language models via bi-modal adversarial prompt}.
arXiv preprint arXiv:2406.04031
(2024)
\end{botherref}
\endbibitem

\bibitem[\protect\citeauthoryear{Ying et~al.}{2024b}]{ying2024unveiling}
\begin{botherref}
\oauthor{\bsnm{Ying}, \binits{Z.}},
\oauthor{\bsnm{Liu}, \binits{A.}},
\oauthor{\bsnm{Liu}, \binits{X.}},
\oauthor{\bsnm{Tao}, \binits{D.}}:
{Unveiling the safety of gpt-4o: An empirical study using jailbreak attacks}.
arXiv preprint arXiv:2406.06302
(2024)
\end{botherref}
\endbibitem

\bibitem[\protect\citeauthoryear{Ying et~al.}{2025}]{ying2025towards}
\begin{botherref}
\oauthor{\bsnm{Ying}, \binits{Z.}},
\oauthor{\bsnm{Zheng}, \binits{G.}},
\oauthor{\bsnm{Huang}, \binits{Y.}},
\oauthor{\bsnm{Zhang}, \binits{D.}},
\oauthor{\bsnm{Zhang}, \binits{W.}},
\oauthor{\bsnm{Zou}, \binits{Q.}},
\oauthor{\bsnm{Liu}, \binits{A.}},
\oauthor{\bsnm{Liu}, \binits{X.}},
\oauthor{\bsnm{Tao}, \binits{D.}}:
{Towards Understanding the Safety Boundaries of DeepSeek Models: Evaluation and Findings}.
arXiv preprint arXiv:2503.15092
(2025)
\end{botherref}
\endbibitem

\bibitem[\protect\citeauthoryear{Jing et~al.}{2025}]{jing2025cogmorph}
\begin{botherref}
\oauthor{\bsnm{Jing}, \binits{Z.}},
\oauthor{\bsnm{Ying}, \binits{Z.}},
\oauthor{\bsnm{Wang}, \binits{L.}},
\oauthor{\bsnm{Liang}, \binits{S.}},
\oauthor{\bsnm{Liu}, \binits{A.}},
\oauthor{\bsnm{Liu}, \binits{X.}},
\oauthor{\bsnm{Tao}, \binits{D.}}:
{CogMorph: Cognitive Morphing Attacks for Text-to-Image Models}.
arXiv preprint arXiv:2501.11815
(2025)
\end{botherref}
\endbibitem

\bibitem[\protect\citeauthoryear{Gan et~al.}{2024}]{Navigate2024}
\begin{botherref}
\oauthor{\bsnm{Gan}, \binits{Y.}},
\oauthor{\bsnm{Yang}, \binits{Y.}},
\oauthor{\bsnm{Ma}, \binits{Z.}},
\oauthor{\bsnm{He}, \binits{P.}},
\oauthor{\bsnm{Zeng}, \binits{R.}},
\oauthor{\bsnm{Wang}, \binits{Y.}},
\oauthor{\bsnm{Li}, \binits{Q.}},
\oauthor{\bsnm{Zhou}, \binits{C.}},
\oauthor{\bsnm{Li}, \binits{S.}},
\oauthor{\bsnm{Wang}, \binits{T.}},
\oauthor{\bsnm{Gao}, \binits{Y.}},
\oauthor{\bsnm{Wu}, \binits{Y.}},
\oauthor{\bsnm{Ji}, \binits{S.}}:
{Navigating the Risks: A Survey of Security, Privacy, and Ethics Threats in LLM-Based Agents}.
arXiv preprint arXiv:2411.09523
(2024)
\end{botherref}
\endbibitem

\bibitem[\protect\citeauthoryear{Liu et~al.}{2024}]{LiuFormalize2024}
\begin{bchapter}
\bauthor{\bsnm{Liu}, \binits{Y.}},
\bauthor{\bsnm{Jia}, \binits{Y.}},
\bauthor{\bsnm{Geng}, \binits{R.}},
\bauthor{\bsnm{Jia}, \binits{J.}},
\bauthor{\bsnm{Gong}, \binits{N.Z.}}:
\bctitle{{Formalizing and Benchmarking Prompt Injection Attacks and Defenses}}.
In: \bbtitle{{USENIX} Security Symposium}
(\byear{2024})
\end{bchapter}
\endbibitem

\bibitem[\protect\citeauthoryear{Zou et~al.}{2023}]{ZouGCG2023}
\begin{botherref}
\oauthor{\bsnm{Zou}, \binits{A.}},
\oauthor{\bsnm{Wang}, \binits{Z.}},
\oauthor{\bsnm{Carlini}, \binits{N.}},
\oauthor{\bsnm{Nasr}, \binits{M.}},
\oauthor{\bsnm{Kolter}, \binits{J.Z.}},
\oauthor{\bsnm{Fredrikson}, \binits{M.}}:
{Universal and Transferable Adversarial Attacks on Aligned Language Models}.
arXiv preprint arXiv:2307.15043
(2023)
\end{botherref}
\endbibitem

\bibitem[\protect\citeauthoryear{Shi et~al.}{2024}]{ShiJudge2024}
\begin{bchapter}
\bauthor{\bsnm{Shi}, \binits{J.}},
\bauthor{\bsnm{Yuan}, \binits{Z.}},
\bauthor{\bsnm{Liu}, \binits{Y.}},
\bauthor{\bsnm{Huang}, \binits{Y.}},
\bauthor{\bsnm{Zhou}, \binits{P.}},
\bauthor{\bsnm{Sun}, \binits{L.}},
\bauthor{\bsnm{Gong}, \binits{N.Z.}}:
\bctitle{{Optimization-based Prompt Injection Attack to LLM-as-a-Judge}}.
In: \bbtitle{Proceedings of the ACM SIGSAC Conference on Computer and Communications Security}
(\byear{2024})
\end{bchapter}
\endbibitem

\bibitem[\protect\citeauthoryear{Gong et~al.}{2025}]{Gongfig2025}
\begin{botherref}
\oauthor{\bsnm{Gong}, \binits{Y.}},
\oauthor{\bsnm{Ran}, \binits{D.}},
\oauthor{\bsnm{Liu}, \binits{J.}},
\oauthor{\bsnm{Wang}, \binits{C.}},
\oauthor{\bsnm{Cong}, \binits{T.}},
\oauthor{\bsnm{Wang}, \binits{A.}},
\oauthor{\bsnm{Duan}, \binits{S.}},
\oauthor{\bsnm{Wang}, \binits{X.}}:
{FigStep: Jailbreaking Large Vision-Language Models via Typographic Visual Prompts}.
arXiv preprint arXiv:2311.05608
(2025)
\end{botherref}
\endbibitem

\bibitem[\protect\citeauthoryear{Kimura et~al.}{2024}]{KimuraHijack2024}
\begin{botherref}
\oauthor{\bsnm{Kimura}, \binits{S.}},
\oauthor{\bsnm{Tanaka}, \binits{R.}},
\oauthor{\bsnm{Miyawaki}, \binits{S.}},
\oauthor{\bsnm{Suzuki}, \binits{J.}},
\oauthor{\bsnm{Sakaguchi}, \binits{K.}}:
{Empirical Analysis of Large Vision-Language Models against Goal Hijacking via Visual Prompt Injection}.
arXiv preprint arXiv:2408.03554
(2024)
\end{botherref}
\endbibitem

\bibitem[\protect\citeauthoryear{Bagdasaryan et~al.}{2023}]{BagdasaryanAbuse2023}
\begin{botherref}
\oauthor{\bsnm{Bagdasaryan}, \binits{E.}},
\oauthor{\bsnm{Hsieh}, \binits{T.-Y.}},
\oauthor{\bsnm{Nassi}, \binits{B.}},
\oauthor{\bsnm{Shmatikov}, \binits{V.}}:
{Abusing Images and Sounds for Indirect Instruction Injection in Multi-Modal LLMs}.
arXiv preprint arXiv:2307.10490
(2023)
\end{botherref}
\endbibitem

\bibitem[\protect\citeauthoryear{Lewis et~al.}{2020}]{LewisRAG2020}
\begin{bchapter}
\bauthor{\bsnm{Lewis}, \binits{P.}},
\bauthor{\bsnm{Perez}, \binits{E.}},
\bauthor{\bsnm{Piktus}, \binits{A.}},
\bauthor{\bsnm{Petroni}, \binits{F.}},
\bauthor{\bsnm{Karpukhin}, \binits{V.}},
\bauthor{\bsnm{Goyal}, \binits{N.}},
\bauthor{\bsnm{Küttler}, \binits{H.}},
\bauthor{\bsnm{Lewis}, \binits{M.}},
\bauthor{\bsnm{Yih}, \binits{W.-t.}},
\bauthor{\bsnm{Rocktäschel}, \binits{T.}},
\bauthor{\bsnm{Riedel}, \binits{S.}},
\bauthor{\bsnm{Kiela}, \binits{D.}}:
\bctitle{{Retrieval-Augmented Generation for Knowledge-Intensive NLP Tasks}}.
In: \bbtitle{Advances in Neural Information Processing Systems}
(\byear{2020})
\end{bchapter}
\endbibitem

\bibitem[\protect\citeauthoryear{{Antropic}}{2024}]{AntropicClaude2024}
\begin{botherref}
\oauthor{\bsnm{{Antropic}}}:
{BUILD WITH CLAUDE: PDF Support}.
\url{https://docs.anthropic.com/en/docs/build-with-claude/pdf-support}
\end{botherref}
\endbibitem

\bibitem[\protect\citeauthoryear{Wallace et~al.}{2024}]{WallaceHierachy2024}
\begin{botherref}
\oauthor{\bsnm{Wallace}, \binits{E.}},
\oauthor{\bsnm{Xiao}, \binits{K.}},
\oauthor{\bsnm{Leike}, \binits{R.}},
\oauthor{\bsnm{Weng}, \binits{L.}},
\oauthor{\bsnm{Heidecke}, \binits{J.}},
\oauthor{\bsnm{Beutel}, \binits{A.}}:
{The Instruction Hierarchy: Training LLMs to Prioritize Privileged Instructions}.
arXiv preprint arXiv:2406.09246
(2024)
\end{botherref}
\endbibitem

\bibitem[\protect\citeauthoryear{OpenAI}{2021}]{chatGPT}
\begin{botherref}
\oauthor{\bsnm{OpenAI}}:
ChatGPT.
\url{https://chatgpt.com}
\end{botherref}
\endbibitem

\bibitem[\protect\citeauthoryear{xAI}{}]{Grok}
\begin{botherref}
\oauthor{\bsnm{xAI}}:
Grok.
\url{https://x.ai}
\end{botherref}
\endbibitem

\bibitem[\protect\citeauthoryear{Cui et~al.}{2024}]{CuiRobust2024}
\begin{bchapter}
\bauthor{\bsnm{Cui}, \binits{X.}},
\bauthor{\bsnm{Aparcedo}, \binits{A.}},
\bauthor{\bsnm{Jang}, \binits{Y.K.}},
\bauthor{\bsnm{Lim}, \binits{S.-N.}}:
\bctitle{{On the Robustness of Large Multimodal Models Against Image Adversarial}}.
In: \bbtitle{Proceedings of the IEEE/CVF International Conference on Computer Vision and Pattern Recognition}
(\byear{2024})
\end{bchapter}
\endbibitem

\bibitem[\protect\citeauthoryear{Liang et~al.}{2022}]{VictorGap2022}
\begin{bchapter}
\bauthor{\bsnm{Liang}, \binits{V.W.}},
\bauthor{\bsnm{Zhang}, \binits{Y.}},
\bauthor{\bsnm{Kwon}, \binits{Y.}},
\bauthor{\bsnm{Yeung}, \binits{S.}},
\bauthor{\bsnm{Zou}, \binits{J.Y.}}:
\bctitle{{Mind the Gap: Understanding the Modality Gap in Multi-modal Contrastive Representation Learning}}.
In: \bbtitle{Advances in Neural Information Processing Systems}
(\byear{2022})
\end{bchapter}
\endbibitem

\bibitem[\protect\citeauthoryear{Zhao et~al.}{2023}]{ZhaoRobust2023}
\begin{bchapter}
\bauthor{\bsnm{Zhao}, \binits{Y.}},
\bauthor{\bsnm{Pang}, \binits{T.}},
\bauthor{\bsnm{Du}, \binits{C.}},
\bauthor{\bsnm{Yang}, \binits{X.}},
\bauthor{\bsnm{LI}, \binits{C.}},
\bauthor{\bsnm{Cheung}, \binits{N.-M.M.}},
\bauthor{\bsnm{Lin}, \binits{M.}}:
\bctitle{{On Evaluating Adversarial Robustness of Large Vision-Language Models}}.
In: \bbtitle{Advances in Neural Information Processing Systems}
(\byear{2023})
\end{bchapter}
\endbibitem

\bibitem[\protect\citeauthoryear{Rombach et~al.}{2022}]{RobinSD2022}
\begin{bchapter}
\bauthor{\bsnm{Rombach}, \binits{R.}},
\bauthor{\bsnm{Blattmann}, \binits{A.}},
\bauthor{\bsnm{Lorenz}, \binits{D.}},
\bauthor{\bsnm{Esser}, \binits{P.}},
\bauthor{},
\bauthor{\bsnm{Ommer}, \binits{B.}}:
\bctitle{{Highresolution image synthesis with latent diffusion models}}.
In: \bbtitle{Proceedings of the IEEE/CVF International Conference on Computer Vision}
(\byear{2022})
\end{bchapter}
\endbibitem

\bibitem[\protect\citeauthoryear{Radford et~al.}{2021}]{RadfordCLIP2021}
\begin{bchapter}
\bauthor{\bsnm{Radford}, \binits{A.}},
\bauthor{\bsnm{Kim}, \binits{J.W.}},
\bauthor{\bsnm{Hallacy}, \binits{C.}},
\bauthor{\bsnm{Ramesh}, \binits{A.}},
\bauthor{\bsnm{Goh}, \binits{G.}},
\bauthor{\bsnm{Agarwal}, \binits{S.}},
\bauthor{\bsnm{Sastry}, \binits{G.}},
\bauthor{\bsnm{Askell}, \binits{A.}},
\bauthor{\bsnm{Mishkin}, \binits{P.}},
\bauthor{\bsnm{Clark}, \binits{J.}},
\bauthor{\bsnm{Krueger}, \binits{G.}},
\bauthor{\bsnm{Sutskever}, \binits{I.}}:
\bctitle{{Learning Transferable Visual Models From Natural Language Supervision}}.
In: \bbtitle{International Conference on Machine Learning}
(\byear{2021})
\end{bchapter}
\endbibitem

\bibitem[\protect\citeauthoryear{Zhai et~al.}{2023}]{ZhaiSigLIP2023}
\begin{bchapter}
\bauthor{\bsnm{Zhai}, \binits{X.}},
\bauthor{\bsnm{Mustafa}, \binits{B.}},
\bauthor{\bsnm{Kolesnikov}, \binits{A.}},
\bauthor{\bsnm{Beyer}, \binits{L.}}:
\bctitle{{Sigmoid Loss for Language Image Pre-Training}}.
In: \bbtitle{International Conference on Computer Vision}
(\byear{2023})
\end{bchapter}
\endbibitem

\bibitem[\protect\citeauthoryear{Chen et~al.}{2024}]{ChenSC2024}
\begin{bchapter}
\bauthor{\bsnm{Chen}, \binits{H.}},
\bauthor{\bsnm{Zhang}, \binits{Y.}},
\bauthor{\bsnm{Dong}, \binits{Y.}},
\bauthor{\bsnm{Yang}, \binits{X.}},
\bauthor{\bsnm{Su}, \binits{H.}},
\bauthor{\bsnm{Zhu}, \binits{J.}}:
\bctitle{Rethinking model ensemble in transfer-based adversarial attacks}.
In: \bbtitle{International Conference on Learning Representations}
(\byear{2024})
\end{bchapter}
\endbibitem

\bibitem[\protect\citeauthoryear{Zhang et~al.}{2025}]{ZhangPoison2025}
\begin{bchapter}
\bauthor{\bsnm{Zhang}, \binits{K.}},
\bauthor{\bsnm{Tao}, \binits{K.}},
\bauthor{\bsnm{Tang}, \binits{J.}},
\bauthor{\bsnm{Wang}, \binits{H.}}:
\bctitle{Poison as cure: Visual noise for mitigating object hallucinations in lvms}.
(\byear{2025})
\end{bchapter}
\endbibitem

\bibitem[\protect\citeauthoryear{OpenAI}{2023}]{GPT4}
\begin{botherref}
\oauthor{\bsnm{OpenAI}}:
{GPT-4 Technical Report}.
arXiv preprint arXiv:2303.08774
(2023)
\end{botherref}
\endbibitem

\bibitem[\protect\citeauthoryear{Zhang et~al.}{2025}]{zhangmeta2023}
\begin{botherref}
\oauthor{\bsnm{Zhang}, \binits{Y.}},
\oauthor{\bsnm{Yuan}, \binits{Y.}},
\oauthor{\bsnm{Yao}, \binits{A.C.-C.}}:
{Meta Prompting for AI Systems}.
arXiv preprint arXiv:2311.11482
(2025)
\end{botherref}
\endbibitem

\bibitem[\protect\citeauthoryear{Wang et~al.}{2024}]{Qwen2VL}
\begin{botherref}
\oauthor{\bsnm{Wang}, \binits{P.}},
\oauthor{\bsnm{Bai}, \binits{S.}},
\oauthor{\bsnm{Tan}, \binits{S.}},
\oauthor{\bsnm{Wang}, \binits{S.}},
\oauthor{\bsnm{Fan}, \binits{Z.}},
\oauthor{\bsnm{Bai}, \binits{J.}},
\oauthor{\bsnm{Chen}, \binits{K.}},
\oauthor{\bsnm{Liu}, \binits{X.}},
\oauthor{\bsnm{Wang}, \binits{J.}},
\oauthor{\bsnm{Ge}, \binits{W.}},
\oauthor{\bsnm{Fan}, \binits{Y.}},
\oauthor{\bsnm{Dang}, \binits{K.}},
\oauthor{\bsnm{Du}, \binits{M.}},
\oauthor{\bsnm{Ren}, \binits{X.}},
\oauthor{\bsnm{Men}, \binits{R.}},
\oauthor{\bsnm{Liu}, \binits{D.}},
\oauthor{\bsnm{Zhou}, \binits{C.}},
\oauthor{\bsnm{Zhou}, \binits{J.}},
\oauthor{\bsnm{Lin}, \binits{J.}}:
{Qwen2-VL: Enhancing Vision-Language Model's Perception of the World at Any Resolution}.
arXiv preprint arXiv:2409.12191
(2024)
\end{botherref}
\endbibitem

\bibitem[\protect\citeauthoryear{{Qwen Team}}{2024}]{Qwen2Blog}
\begin{botherref}
\oauthor{\bsnm{{Qwen Team}}}:
{Qwen2-VL: To See the World More Clearly}.
\url{https://qwenlm.github.io/blog/qwen2-vl/}
\end{botherref}
\endbibitem

\bibitem[\protect\citeauthoryear{Ying et~al.}{2024}]{YingSafe2024}
\begin{botherref}
\oauthor{\bsnm{Ying}, \binits{Z.}},
\oauthor{\bsnm{Liu}, \binits{A.}},
\oauthor{\bsnm{Liang}, \binits{S.}},
\oauthor{\bsnm{Huang}, \binits{L.}},
\oauthor{\bsnm{Guo}, \binits{J.}},
\oauthor{\bsnm{Zhou}, \binits{W.}},
\oauthor{\bsnm{Liu}, \binits{X.}},
\oauthor{\bsnm{Tao}, \binits{D.}}:
{SafeBench: A Safety Evaluation Framework for Multimodal Large Language Models}.
arXiv preprint arXiv:2410.18927
(2024)
\end{botherref}
\endbibitem

\bibitem[\protect\citeauthoryear{Raheja et~al.}{2023}]{raheja2023coedit}
\begin{botherref}
\oauthor{\bsnm{Raheja}, \binits{V.}},
\oauthor{\bsnm{Kumar}, \binits{D.}},
\oauthor{\bsnm{Koo}, \binits{R.}},
\oauthor{\bsnm{Kang}, \binits{D.}}:
{CoEDIT: Text Editing by Task-Specific Instruction Tuning}.
arXiv preprint arXiv:2305.09857
(2023)
\end{botherref}
\endbibitem

\bibitem[\protect\citeauthoryear{Socher et~al.}{2013}]{SocherRecursive2013}
\begin{bchapter}
\bauthor{\bsnm{Socher}, \binits{R.}},
\bauthor{\bsnm{Perelygin}, \binits{A.}},
\bauthor{\bsnm{Wu}, \binits{J.}},
\bauthor{\bsnm{Chuang}, \binits{J.}},
\bauthor{\bsnm{Manning}, \binits{C.D.}},
\bauthor{\bsnm{Ng}, \binits{A.}},
\bauthor{\bsnm{Potts}, \binits{C.}}:
\bctitle{{Recursive Deep Models for Semantic Compositionality Over a Sentiment Treebank}}.
In: \bbtitle{Conference on Empirical Methods in Natural Language Processing}
(\byear{2013})
\end{bchapter}
\endbibitem

\bibitem[\protect\citeauthoryear{Husain et~al.}{2020}]{HusainCodesearchnet2020}
\begin{botherref}
\oauthor{\bsnm{Husain}, \binits{H.}},
\oauthor{\bsnm{Wu}, \binits{H.-H.}},
\oauthor{\bsnm{Gazit}, \binits{T.}},
\oauthor{\bsnm{Allamanis}, \binits{M.}},
\oauthor{\bsnm{Brockschmidt}, \binits{M.}}:
{CodeSearchNet Challenge: Evaluating the State of Semantic Code Search}.
arXiv preprint arXiv:1909.09436
(2020)
\end{botherref}
\endbibitem

\bibitem[\protect\citeauthoryear{{W3C}}{2011}]{HTML5}
\begin{botherref}
\oauthor{\bsnm{{W3C}}}:
{HTML5}.
\url{https://www.w3.org/TR/2011/WD-html5-20110405/}
\end{botherref}
\endbibitem

\bibitem[\protect\citeauthoryear{Li et~al.}{2024}]{Lillmsasjudges2024}
\begin{botherref}
\oauthor{\bsnm{Li}, \binits{H.}},
\oauthor{\bsnm{Dong}, \binits{Q.}},
\oauthor{\bsnm{Chen}, \binits{J.}},
\oauthor{\bsnm{Su}, \binits{H.}},
\oauthor{\bsnm{Zhou}, \binits{Y.}},
\oauthor{\bsnm{Ai}, \binits{Q.}},
\oauthor{\bsnm{Ye}, \binits{Z.}},
\oauthor{\bsnm{Liu}, \binits{Y.}}:
{LLMs-as-Judges: A Comprehensive Survey on LLM-based Evaluation Methods}.
arXiv preprint arXiv:2412.05579
(2024)
\end{botherref}
\endbibitem

\bibitem[\protect\citeauthoryear{Carlini et~al.}{2019}]{CarliniAdversarial2019}
\begin{botherref}
\oauthor{\bsnm{Carlini}, \binits{N.}},
\oauthor{\bsnm{Athalye}, \binits{A.}},
\oauthor{\bsnm{Papernot}, \binits{N.}},
\oauthor{\bsnm{Brendel}, \binits{W.}},
\oauthor{\bsnm{Rauber}, \binits{J.}},
\oauthor{\bsnm{Tsipras}, \binits{D.}},
\oauthor{\bsnm{Goodfellow}, \binits{I.}},
\oauthor{\bsnm{Madry}, \binits{A.}},
\oauthor{\bsnm{Kurakin}, \binits{A.}}:
{On Evaluating Adversarial Robustness}.
arXiv preprint arXiv:1902.06705
(2019)
\end{botherref}
\endbibitem

\bibitem[\protect\citeauthoryear{{Qwen Team}}{2024}]{qwen25}
\begin{botherref}
\oauthor{\bsnm{{Qwen Team}}}:
{Qwen2.5 technical report}.
arXiv preprint arXiv:2412.15115
(2024)
\end{botherref}
\endbibitem

\bibitem[\protect\citeauthoryear{{The Vicuna Team}}{{2023}}]{vicuna}
\begin{botherref}
\oauthor{\bsnm{{The Vicuna Team}}}:
{Vicuna: An Open-Source Chatbot Impressing GPT-4 with 90\% ChatGPT Quality}.
\url{{https://lmsys.org/blog/2023-03-30-vicuna/}}
\end{botherref}
\endbibitem

\bibitem[\protect\citeauthoryear{{Mistral AI team}}{2023}]{mistral}
\begin{botherref}
\oauthor{\bsnm{{Mistral AI team}}}:
{Mistral 7B}.
\url{https://mistral.ai/news/announcing-mistral-7b}
\end{botherref}
\endbibitem

\bibitem[\protect\citeauthoryear{Shazeer et~al.}{2017}]{shazeerMoE2017}
\begin{bchapter}
\bauthor{\bsnm{Shazeer}, \binits{N.}},
\bauthor{\bsnm{Mirhoseini}, \binits{A.}},
\bauthor{\bsnm{Maziarz}, \binits{K.}},
\bauthor{\bsnm{Davis}, \binits{A.}},
\bauthor{\bsnm{Le}, \binits{Q.}},
\bauthor{\bsnm{Hinton}, \binits{G.}},
\bauthor{\bsnm{Dean}, \binits{J.}}:
\bctitle{{Outrageously Large Neural Networks: The Sparsely-Gated Mixture-of-Experts Layer}}.
In: \bbtitle{{International Conference on Learning Representations}}
(\byear{2017})
\end{bchapter}
\endbibitem

\bibitem[\protect\citeauthoryear{Robotics}{2025}]{limo}
\begin{botherref}
\oauthor{\bsnm{Robotics}, \binits{A.}}:
{AgileX Robotics}.
\url{https://global.agilex.ai/pages/limo}
\end{botherref}
\endbibitem

\bibitem[\protect\citeauthoryear{Liu et~al.}{2019}]{liuperceptual2019}
\begin{bchapter}
\bauthor{\bsnm{Liu}, \binits{A.}},
\bauthor{\bsnm{Liu}, \binits{X.}},
\bauthor{\bsnm{Fan}, \binits{J.}},
\bauthor{\bsnm{Ma}, \binits{Y.}},
\bauthor{\bsnm{Zhang}, \binits{A.}},
\bauthor{\bsnm{Xie}, \binits{H.}},
\bauthor{\bsnm{Tao}, \binits{D.}}:
\bctitle{{Perceptual-sensitive gan for generating adversarial patches}}.
In: \bbtitle{Proceedings of the AAAI Conference on Artificial Intelligence}
(\byear{2019})
\end{bchapter}
\endbibitem

\bibitem[\protect\citeauthoryear{Schulhoff}{2024}]{Sandwich}
\begin{botherref}
\oauthor{\bsnm{Schulhoff}, \binits{S.}}:
{Sandwitch Defense.}
\url{https://learnprompting.org/docs/prompt_hacking/defensive_measures/sandwich_defense}
\end{botherref}
\endbibitem

\bibitem[\protect\citeauthoryear{Zhang et~al.}{2025}]{ZhangJailGuard2025}
\begin{botherref}
\oauthor{\bsnm{Zhang}, \binits{X.}},
\oauthor{\bsnm{Zhang}, \binits{C.}},
\oauthor{\bsnm{Li}, \binits{T.}},
\oauthor{\bsnm{Huang}, \binits{Y.}},
\oauthor{\bsnm{Jia}, \binits{X.}},
\oauthor{\bsnm{Hu}, \binits{M.}},
\oauthor{\bsnm{Zhang}, \binits{J.}},
\oauthor{\bsnm{Liu}, \binits{Y.}},
\oauthor{\bsnm{Ma}, \binits{S.}},
\oauthor{\bsnm{Shen}, \binits{C.}}:
{JailGuard: A Universal Detection Framework for LLM Prompt-based Attacks}.
arXiv preprint arXiv:2312.10766
(2025)
\end{botherref}
\endbibitem

\bibitem[\protect\citeauthoryear{Liu et~al.}{2023}]{liu2023towards}
\begin{botherref}
\oauthor{\bsnm{Liu}, \binits{A.}},
\oauthor{\bsnm{Tang}, \binits{S.}},
\oauthor{\bsnm{Chen}, \binits{X.}},
\oauthor{\bsnm{Huang}, \binits{L.}},
\oauthor{\bsnm{Qin}, \binits{H.}},
\oauthor{\bsnm{Liu}, \binits{X.}},
\oauthor{\bsnm{Tao}, \binits{D.}}:
{Towards Defending Multiple Lp-norm Bounded Adversarial Perturbations via Gated Batch Normalization}.
International Journal of Computer Vision
(2023)
\end{botherref}
\endbibitem

\bibitem[\protect\citeauthoryear{Zhang et~al.}{2021}]{zhang2021interpreting}
\begin{botherref}
\oauthor{\bsnm{Zhang}, \binits{C.}},
\oauthor{\bsnm{Liu}, \binits{A.}},
\oauthor{\bsnm{Liu}, \binits{X.}},
\oauthor{\bsnm{Xu}, \binits{Y.}},
\oauthor{\bsnm{Yu}, \binits{H.}},
\oauthor{\bsnm{Ma}, \binits{Y.}},
\oauthor{\bsnm{Li}, \binits{T.}}:
{Interpreting and Improving Adversarial Robustness of Deep Neural Networks with Neuron Sensitivity}.
IEEE Transactions on Image Processing
(2021)
\end{botherref}
\endbibitem

\bibitem[\protect\citeauthoryear{Liu et~al.}{2021}]{liu2021training}
\begin{botherref}
\oauthor{\bsnm{Liu}, \binits{A.}},
\oauthor{\bsnm{Liu}, \binits{X.}},
\oauthor{\bsnm{Yu}, \binits{H.}},
\oauthor{\bsnm{Zhang}, \binits{C.}},
\oauthor{\bsnm{Liu}, \binits{Q.}},
\oauthor{\bsnm{Tao}, \binits{D.}}:
{Training robust deep neural networks via adversarial noise propagation}.
TIP
(2021)
\end{botherref}
\endbibitem

\end{thebibliography}
